\documentclass{article}

\usepackage{PRIMEarxiv}

\usepackage[utf8]{inputenc} 
\usepackage[T1]{fontenc} 
\usepackage{float}
\usepackage{hyperref}  
\usepackage{url}  
\usepackage{booktabs}    
\usepackage{nicefrac} 
\usepackage{microtype} 
\usepackage{lipsum}
\usepackage{fancyhdr} 
\usepackage{graphicx} 
\graphicspath{{media/}} 
\usepackage{adjustbox}  
\usepackage[numbers]{natbib}

\usepackage{multirow}%
\usepackage{amsmath,amssymb,amsfonts}%
\usepackage{amsthm}%
\usepackage{mathrsfs}%
\usepackage[title]{appendix}%
\usepackage{xcolor}%
\usepackage{textcomp}%
\usepackage{manyfoot}%
\usepackage{algorithm}%
\usepackage{algorithmicx}%
\usepackage{algpseudocode}%
\usepackage{listings}%
\usepackage{threeparttable}
\usepackage{array}
\usepackage{tikz}
\usepackage{physics}
\usepackage{mathdots}
\usepackage{yhmath}
\usepackage{cancel}
\usepackage{color}
\usepackage{siunitx}
\usepackage{gensymb}
\usepackage{tabularx}
\usepackage{extarrows}
\usetikzlibrary{fadings}
\usetikzlibrary{patterns}
\usetikzlibrary{shadows.blur}
\usetikzlibrary{shapes}
\usepackage{subcaption}

\usepackage{natbib}

\usepackage[scaled]{helvet} % for sans-serif text
 
\usepackage{mathpazo} % for Palatino-like math fonts
\title{\textbf{FlightScope}: An Experimental Comparative Review of Aircraft Detection Algorithms in Satellite Imagery}

\author{
  Safouane EL GHAZOUALI*\\
   Machine Learning Research\\\& Development, TOELT LLC AI lab\\
  Winterthur, Switzerland\\
  \texttt{safouane.elghazouali@toelt.ai} \\
  \And
  Arnaud GUCCIARDI\\
   Machine Learning Research \\\& Development, TOELT LLC AI lab\\
  Winterthur, Switzerland \\
  \texttt{arnaud.gucciardi@toelt.ai}\\
  \And
  Francesca VENTURINI\\
  Machine Learning Research \\\& Development, TOELT LLC AI lab\\ Winterthur, Switzerland \\
  Institute of Applied Mathematics \\
  \& Physics,  ZHAW - Zurich \\ University of Applied Sciences, \\
  Winterthur, Switzerland \\
  \texttt{vent@zhaw.ch} /\\
  \texttt{francesca.venturini@toelt.ai} \\
  \And
  Nicola VENTURI\\
  Competence Center for AI\\
  and Simulation, armasuisse S+T, \\
  Thun, Switzerland \\
  \texttt{nicola.venturi@armasuisse.ch}\\
  \And
  Michael RÜEGSEGGER\\
  Competence Center for AI\\
  and Simulation, armasuisse S+T, \\
  Thun, Switzerland \\
  \texttt{michael.rueegsegger@armasuisse.ch}\\
  \And
  Umberto MICHELUCCI \\
  Machine Learning Research \\\& Development, TOELT LLC AI lab\\ Winterthur, Switzerland \\
  Computer Science Department,\\ Lucerne University of Applied \\Science and Arts, \\Luzern, Switzerland \\
  \texttt{umberto.michelucci@toelt.ai} / \\ \texttt{umberto.michelucci@hslu.ch} \\
}

\begin{document}
\maketitle

\begin{abstract}
    Object detection in remotely sensed satellite pictures is fundamental in many fields such as biophysical, and environmental monitoring. While deep learning algorithms are constantly evolving, they have been mostly implemented and tested on popular ground-based taken photos. This paper critically evaluates and compares a suite of advanced object detection algorithms customized for the task of identifying aircraft within satellite imagery. Using the large HRPlanesV2 dataset, together with a rigorous validation with the GDIT dataset, this research encompasses an array of methodologies including YOLO versions 5 and 8, Faster RCNN, CenterNet, RetinaNet, RTMDet, and DETR, all trained from scratch. This exhaustive training and validation study reveal YOLOv5 as the preeminent model for the specific case of identifying airplanes from remote sensing data, showcasing high precision and adaptability across diverse imaging conditions. This research highlight the nuanced performance landscapes of these algorithms, with YOLOv5 emerging as a robust solution for aerial object detection, underlining its importance through superior mean average precision, Recall, and Intersection over Union scores. The findings described here underscore the fundamental role of algorithm selection aligned with the specific demands of satellite imagery analysis and extend a comprehensive framework to evaluate model efficacy. The benchmark toolkit and codes, available via  \href{https://github.com/toelt-llc/FlightScope\_Bench}{GitHub}, aims to further exploration and innovation in the realm of remote sensing object detection, paving the way for improved analytical methodologies in satellite imagery applications.
\end{abstract}

% keywords can be removed
\keywords{Object detection \and survey \and remote sensing \and satellite image \and aircraft localization}

\section{Introduction}\label{sec1}

Remote sensing plays a fundamental role in acquiring information about the Earth's surface using various types of vision sensors such as Thermal Infra-Red (TIR), synthetic aperture radar (SAR), inverse SAR (ISAR) \cite{zhang2023application} and RGB cameras \cite{ashapure2019comparative, daudt2023weakly}. This field encompasses a wide range of technologies and methodologies that aim to capture, analyze, and interpret data from various sources. Within the realm of remote sensing, one of the most significant applications is the detection and localization of small objects \cite{fu2023anchor}. Object detection from satellite imagery is of great importance in various domains, such as defense and military applications \cite{9821591}, urban studies \cite{10.1007/978-3-540-69839-5_23}, earth monitoring and assessment \cite{najafzadeh2021reliability}, airport surveillance and georeferencing \cite{sohl2024low}, vessel traffic monitoring \cite{greidanus2008satellite} and transportation infrastructure determination \cite{liu2022aircraft, blattner2021commercial}. Unlike photographic images, remote sensing images obtained from satellite sensors are more difficult to interpret due to factors such as atmospheric interference, viewpoint variation, background clutter, and illumination differences \cite{prudyus2015factors, rs12030458}. \linebreak  Additionally, satellite images cover larger areas, typically 10 $\times$ 10 km$^2$ per frame, %EE: check meaning retained
representing the intricate landscape of the Earth's surface with two-dimensional images that possess less spatial detail compared to digital photographs from cameras.

Traditional approaches to aircraft detection relied on manual feature engineering and machine learning techniques. However, these methods often struggle to handle the complexities of satellite imagery and achieve high accuracy. The advent of deep learning, particularly Convolutional Neural Networks (CNNs), has revolutionized object detection tasks by enabling the automatic extraction of intricate visual representations.
One notable deep learning method for object detection is You Only Look Once (YOLO) \cite{yolov8_ultralytics}. YOLO divides the input image into a grid and predicts bounding boxes and class probabilities directly from the grid cells. Its architecture allows real-time detection and can handle multiple object classes simultaneously. However, YOLO may encounter challenges in accurately localizing small objects due to the grid cell structure and limited receptive fields.
Another popular deep learning technique is Single-Shot MultiBox Detection (SSD) \cite{liu_ssd_2016}, which utilizes a series of convolutional layers to generate a diverse set of default bounding boxes at different scales and aspect ratios. By applying a set of predefined anchor boxes to the feature maps, SSD performs multi-scale object detection efficiently. However, it may face difficulties in detecting small objects and suffers from a large number of default boxes, leading to computational overhead.
Region-based CNNs (RCNN) \cite{girshick_region-based_2016} and their variants, such as Fast RCNN \cite{girshick_fast_2015}, Faster RCNN \cite{ren_faster_2015}, and Mask RCNN \cite{he_mask_2017}, have also been widely used for object detection tasks. These methods employ a two-stage approach in which potential object regions are first proposed and then classified and refined. Using region proposals, these methods achieve accurate localization and exhibit strong performance. However, they are computationally expensive and slower than one-stage detectors such as YOLO and SSD.
In addition to the YOLO, SSD and RCNN variants, there are several other deep learning methods that have been explored for object detection, such as EfficientDet \cite{tan_efficientdet_2020}, RetinaNet \cite{lin_focal_2020}, and CenterNet \cite{duan_centernet_2019}. More recently, the success of the attention mechanism in the field of language processing has received much consideration and has also been brought into the field of computer vision \cite{guo2022attention}. This rise has led to an interesting improvement in performance in many subfields of computer vision, including image classification \cite{meng2023few} and object detection \cite{li2020object}. Each of these methods has its own unique architecture and advantages, aiming to improve both accuracy and efficiency in object detection tasks.

In particular, the objective of this study is to benchmark and compare multiple state-of-the-art object detection methods prepared and trained specifically for the use case of aircraft detection in satellite images, more specifically optical RGB remotely sensed photos. {The main focus is to train advanced predictive models and identify highly relevant models for applications} in satellite surveillance and air traffic management. This paper significantly contributes to research on satellite imagery analysis by implementing, training, and validating  the ten leading object detection neural network architectures listed in Table \ref{tab:models_descriptions}, \linebreak  along with other cutting-edge deep learning architectures. The authors have directly implemented and tested each network to control the testing conditions and prevent bias or spurious outcomes that might arise. 
The algorithms have been chosen according to two criteria: (i) reference architectures have been selected such as RCNN, YOLO, and more recently DETR as they constitute reference bases for comparisons; (ii) the algorithm must have demonstrated previous state-of-the-art performances on benchmark datasets such as COCO \cite{lin2015microsoftcococommonobjects} or ImageNet \cite{5206848}.

\begin{table}[h!]
\renewcommand{\arraystretch}{1.3}
    \centering
    \caption{Overview of the different neural network architectures implemented, trained and validated in this paper.}
    \begin{tabular}{p{0.2\linewidth}p{0.12\linewidth}p{0.70\linewidth}} % Adjusted column widths
        \hline
        \textbf{Models Tested} & \textbf{{Category}} & \textbf{Short Description} \\
        \hline
        {YOLO v10} \cite{THU_MIGyolov10} & {Transformer} & {Leverages Transformer-based modules for improved attention and introduces a novel NMS-free strategy for better efficiency. It combines advanced architectural optimization to reduce computational redundancy while maintaining high detection accuracy}. \\
        YOLO v8 \cite{yolov8_ultralytics} & {One-stage} & Updated version of the YOLO object detection system, incorporating advancements in network architecture and training techniques to achieve better efficiency in real-time object detection tasks. \\
        YOLO v5 \cite{jocher2023yolo} & {One-stage} & Real-time object detection model that processes images in one pass through a CNN, widely used for various applications including autonomous vehicles, surveillance systems, and robotics. \\
        RTMDet \cite{lyu2022RTMDet} & {One-stage} & An improved CNN network for higher accuracy while maintaining the same real-time performance of the YOLO model series. \\
        DETR \cite{lv2023detrs} & {Transformer} & New object detection architecture based on Transformers and attention mechanisms. It has proved its efficiency on the COCO dataset. \\
        Faster RCNN \cite{ren_faster_2015} & {Two-stage} & Region-based CNN is an architecture for object detection that has proven its efficiency in accurate bounding box extraction. Faster RCNN is an algorithm that improves the detection speed of RCNNs.\\
        SSD \cite{liu_ssd_2016} & {One-stage} & Another real-time object detection model that predicts bounding boxes from feature maps at multiple scales. \\
        CenterNet \cite{duan_centernet_2019} & {One-stage} & This model, initially based on the CornerNet \cite{DBLP:CornerNet_journals/corr/abs-1808-01244} architecture, has been able to achieve state-of-the-art performance on the COCO dataset for object localization in term of real-time and accuracy. \\
        RetinaNet \cite{RetinaNet_Lin} & {One-stage} & Real-time object detection model that addresses the imbalance between foreground and background examples during training using a novel focal loss function. \\
        {Grounding DINO} \cite{jocher2023yolo} & Transformer & Cutting-edge Transformer-based open-world object detection model that blends the capabilities of vision Transformers with language understanding. This model excels specifically at zero-shot object detection, enabling it to identify objects not present in its training set using natural \linebreak  language queries. \\
        \hline
    \end{tabular}
    \label{tab:models_descriptions}
\end{table}

In addition to a detailed overview of object detection models that is fundamental for both researchers and practitioners, this work also provides a thorough examination of their performance, precision, and computational complexity. The findings provide critical information that improves the selection process for the most efficient aircraft detection methods in satellite imagery. This is supported by comprehensive training and initial validation on the HRPlanesv2 and further validation on GDIT datasets (for details on the datasets, see Section \ref{sec:datasets_section}), marking an important advancement in the study of the precision and efficiency of remote sensing technologies and their application in real-world scenarios, thus underlining their substantial relevance and potential impact on future research in the field.

Additionally, to make a more useful contribution to the computer vision community, all the code used for the benchmarks in this paper is available and can be reproduced from the GitHub repository at \url{https://github.com/toelt-llc/FlightScope_Bench} (accessed on 23 October 2024).

Following the introduction, the rest of this paper is structured as follows: Similar comparative studies are discussed and reviewed in Section \ref{sec:related_work}. An overview of existing datasets including airplanes in satellite images is discussed in Section \ref{sec:datasets_section}, while in Section \ref{sec:OD_survey} a more detailed state-of-the-art description of object detection architectures (listed in Table \ref{tab:models_descriptions}) is given. Section \ref{sec:result1_bench} presents the setup of the benchmark, along with the results and discussion. Finally, Section \ref{sec:conclusion} concludes and summarizes the findings of this comparative study.

%%%%%%%%%%%%%%%%%%%%%%%%%%%%%%%%%%%%%%%%%%%%%%%
\section{Related work}\label{sec:related_work}

Many remote sensing methods and models have been studied and proposed during the past decade in various fields such as environmental monitoring \cite{Young2017}, object and image geolocalization \cite{wilson2023image, wosner2021object}, urban planning \cite{zhang2023stp}, and agriculture \cite{wosner2021object}. In the context of object detection, the methods that are usually proposed are trained, validated, and tested on images gathered from ground and usual vision sensors. {Object detection and segmentation techniques demonstrate accurate predictive performance in these types of images. Consequently, there is a significant need to evaluate the performance of remote sensing images using these methods}. Remote sensing imagery typically encompasses vast areas with varying resolutions, making the detection of small objects, such as vehicles or infrastructure, particularly challenging \cite{ayush2022vehicle}. In addition, factors such as varying illumination conditions, occlusions due to weather, and object scale variations further complicate the task.

A few surveys on small object detection from aerial optical imagery (summary in Table \ref{tab:object_detection_summary}) have been proposed, such as ship detection \cite{er2023ship}, where the authors conduct a critical review of deep learning networks and available datasets. Another paper by W. Chen 
et al. \cite{rs14215460} compared multiple families of RCNN, YOLO, SSD, CornerNet, and other models on their own dataset HRSC2016\-MS and demonstrated the generalization effectiveness of their model MSSDet based on a ResNet backbone \cite{he2015deepresiduallearningimage}. Ships are relatively easier to detect because of the background aspect of those objects, as in most cases they are found on water, which can be distinguished accurately based on the colors and texture uniformity. A more similar study to this work, conducted by Alganci et al. \cite{rs12030458}, delves into the detection of small objects from satellite imagery. This research focuses on evaluating the performance of three state-of-the-art Convolutional Neural Network (CNN)-based object detection models specifically tailored for identifying airplanes in very-high-resolution (VHR) satellite images.
The authors underscore the importance of accurate and efficient detection methods in satellite imagery due to its large data size and expansive aerial coverage. In another study, %EE: check meaning retained
the authors used the DOTA dataset \cite{xia2018dota}, a multiple-class open-source repository explicitly created for object detection in remote sensing images. This dataset encompasses satellite image patches sourced from platforms such as Google Earth, Jilin 1 (JL-1), and Gaofen 2 (GF-2) satellites, featuring 15 object categories, among which is the airplane class.
The comparative evaluation conducted in \cite{rs12030458} assesses three object detection models, Faster R-CNN \cite{ren_faster_2015}, SSD \cite{liu_ssd_2016}, and YOLO-v3 \cite{redmon2018yolov3}, using the DOTA dataset for training and testing. Performance metrics, including COCO metrics, F1 scores, and processing time, are used for evaluation. The summary of this work reveals that Faster R-CNN exhibits superior detection accuracy, with YOLO-v3 showcasing faster convergence capabilities. SSD, although proficient in object localization, faces challenges with training convergence.

Furthermore, another study presented in \cite{article_1107890} focuses on the development and evaluation of the first version of the HRPlanes benchmark dataset for deep learning-based airplane detection using satellite imagery from Google Earth. The authors describe the HRPlanes dataset and some of the images captured by different satellites to represent diverse landscapes, seasonal variations, and satellite geometry conditions. The dataset is then selected for the training and validation of two widely used object detection methods, YOLOv4 \cite{alexey2020yolov4} and Faster R-CNN \cite{ren_faster_2015}. 
The comparative study between YOLOv4 and Faster R-CNN in the context of airplane detection from satellite imagery reveals interesting findings. The study highlights that the boundaries of bounding boxes for YOLOv4 are better at certain scales compared to Faster R-CNN. For example, in some cases, YOLOv4 performs better in detecting small airplanes, while Faster R-CNN excels in detecting larger ones. The results also indicate that YOLOv4 is more effective in creating accurate bounding boxes for commercial planes in large-scale imagery, possibly due to the presence of boarding bridges near the planes. Additionally, both deep learning models demonstrate the ability to detect moving planes, even in scenarios with motion blur effects in the images.

\begin{table}[H]
\renewcommand{\arraystretch}{1.3}
    \centering
    \caption{Summary of some reviews of optical remote sensing object detection and classification.}
    \begin{tabular} {p{0.2\linewidth}p{0.70\linewidth}p{0.1\linewidth}p{0.1\linewidth}}
        \textbf{Study Description} & \textbf{Summary} & \textbf{Dataset} & \textbf{Classes} \\
        \hline
        Vehicle and Vessel Detection \cite{rs12071217} & Investigates the effectiveness of single-shot object detection networks for identifying small objects like vehicles and vessels in satellite imagery with resolutions of 0.3–0.5 m. Key challenges addressed include small object sizes (ranging from 5 pixels to several hundred) and diverse object dimensions. The study annotated 1500 km\textsuperscript{2} of imagery, equally split between land and water areas. Four models were compared: YOLOv2, YOLOv3, D-YOLO, and YOLT. Extensive hyperparameter tuning was conducted to maximize performance. D-YOLO emerged as the top performer, achieving an average precision (AP) of 60\% for vehicles and 66\% for vessels. YOLOv3 provided balanced results across metrics, while other models showed varying strengths in detection speed and accuracy.
        & DOTA, OpenSARShip & Vehicles, Vessels \\ 
        \hline
        Multi-Scale Ship Detection \cite{rs14215460} & Introduces a framework specifically designed to address the challenges of detecting ships at varying scales in optical remote sensing imagery. Traditional detection methods often struggle with the complexities posed by scale variations, cluttered backgrounds, and low contrast between ships and their surroundings. MSSDet addresses these issues by incorporating a Joint Recursive Feature Pyramid (JRFP) into its architecture, which enhances feature fusion across multiple scales. MSSDet demonstrates robust performance when tested on HRSC2016-MS, the original HRSC2016, and the DIOR dataset, achieving notable mAP improvements over other models like the R-CNN family, YOLOv3, and SSD.
        & HRSC2016, DOTA & Ships \\
        \hline
        Airplane Detection from VHR Images \cite{rs12030458} & Conducts a comprehensive evaluation of popular CNN-based models—YOLOv4, Faster R-CNN, and SSD—focusing on their performance in detecting airplanes from satellite imagery. The models were trained on the DOTA dataset and tested on both the DOTA and Pleiades satellite images. The study found that Faster R-CNN performed the best in terms of COCO metrics and F1 scores, excelling in object localization, particularly for larger airplanes. YOLOv4 demonstrated faster convergence and was more efficient in detecting smaller airplanes, making it suitable for real-time applications where processing speed is crucial. However, SSD, while less effective in detection performance, performed relatively well in object localization. 
        & DOTA, HRPlanes & Airplanes \\ 
        \hline
        Ship Detection Using Wavelet Transforms \cite{rs9100985} & Focuses on improving ship detection and segmentation accuracy in optical remote sensing images by addressing common challenges such as background clutter, environmental noise, and false alarms. The authors propose an approach using wavelet transform to reduce the amount of false positives in the prediction, particularly in cluttered coastal areas. By incorporating multi-level false alarm identification, the method enhances the detection of ships within various challenging environments, such as cloud cover or water surface reflections.
        & Custom dataset from coastal regions & Ships \\
        \hline
        Small Objects in remote sensing \cite{rs15133265} & Focuses on the challenges and methods associated with detecting small objects in remote sensing imagery. It discusses the application of deep learning techniques to improve detection accuracy, especially for small-scale objects like vehicles, buildings, and other infrastructures in high-resolution satellite and aerial imagery. The review highlights the significance of using advanced architectures, such as YOLOv8 and Transformers. Various strategies are also subject of discussion such data augmentation and transfer learning. The authors also conducted a comparative study that includes 3 categories: one-stage, two-stage, and anchor-free. 
        & DOTA & Multiclass: airplane, buildings, vehicles, etc. \\
        \hline
        \end{tabular}
    \label{tab:object_detection_summary}
\end{table}

%%%%%%%%%%%%%%%%%%%%%%%%%%%%%%%%%%%%%%%%%%%%%%%
\section{Aircraft Datasets}\label{sec:datasets_section}
In the field of aircraft detection and remote sensing, access to high-quality and diverse datasets is important for the development and evaluation of computer vision algorithms. In some comparative studies, the DOTA dataset \cite{xia2018dota} is selected,  encompassing objects other than aircraft such as airports, bridges, and containers. This section reviews a collection of datasets specifically and exclusively designed for aircraft detection research, each offering unique features and advantages. An overview of these datasets is given in Table \ref{tab:dataset_overview}.

\begin{table}[h!]
    \centering
    \caption{Overview of the available datasets exclusively created for aircraft detection from aerial imagery.}
    \begin{tabular}{p{0.2\linewidth}p{0.6\linewidth}p{0.1\linewidth}}%p{0.2\linewidth}} % Adjusted column widths
        \hline
        \textbf{Dataset} & \textbf{Short Description} %& \textbf{Resolution} 
        & \textbf{Number of Images} \\
        \hline
        Airbus aircraft \cite{airbus-aircraft-detection_dataset} & The Airbus Aircraft Dataset is extracted from a larger deep learning dataset, created with the use of Airbus satellite imagery. The dataset draws its primary imagery from the Pleiades twin satellites operated by Airbus. Images have a resolution of approximately 50 cm per pixel, stored as JPEG files. These images have a resolution of 2560 x 2560 pixels, representing an on-ground area of 1280 metres.  & 109 \\
        HRPlanesV2 \cite{article_1107890} & The HRPlanesv2 dataset is comprised of 2,120 ultra-high-resolution images from Google Earth, featuring a total of 14,335 labelled aircrafts.
        Each image is preserved in JPEG format, measuring 4800 x 2703 pixels, and the labels for each aircraft are documented in the YOLO text format.  & 2120 \\
        RarePlanes \cite{DBLP:journals/corr/abs-2006-02963} & This dataset incorporates real and synthetically generated satellite images. The `real' portion of the dataset consists of 253 Maxar WorldView-3 satellite scenes, including 112 locations and 2142 km$^2$ with 14700 hand-annotated aircrafts. The `synthetic' portion features 50000 synthetic satellite images with roughly 63000 aircraft annotations. Only the `real' part was used for this paper. & 253 \\
        GDIT \cite{GDIT-airport_dataset} & The GDIT Aerial Airport dataset is composed of aerial photographs that show parked airplanes. All varieties of plane are classified under a single category called `airplane.'  & 338 \\
        Planesnet \cite{robert_hammell_2023} & Planesnet is a collection of images extracted from the Planet satellite imagery. The main purpose of the dataset is the classification and localization of airplanes in medium-resolution images. The dataset includes 32000 very small images (20x20 pixels) labelled with either a "plane" or "no-plane" class. & 32000 \\
        Flying Airpl. \cite{3mbt-tb11-21} & Flying Airplanes is a massive dataset that contains satellite images of flying airplanes that surround 30 different European airports. Images are from the Sentinel-2 satellite. & Not available. \\
        OPT-Aircraft \cite{chen2020dataset} & This dataset is a public remote sensing dataset with images stored in .png format that consists of 3594 data files with an approximate size of 69.3 MB. This dataset allows the identification of aircraft and classifies them according to their type and shape. & 3595 \\
        \hline
    \end{tabular}
    \label{tab:dataset_overview}
\end{table}

%\begin{figure}[h]
%        \centering
%        \includegraphics[width=90mm,scale=1]{dataset_overview2.pdf}%{dataset_overview.pdf}
%        \caption{Datasets overview for aircraft detection datasets. The diagram is limited to 5 out of 7 datasets that contain full High-resolution images only. In addition, only the real part was considered for RarePlanes dataset. }
%        \label{fig:dataset_classification}
%\end{figure}

\subsection{Airbus Aircraft Dataset}

The \textbf{Airbus Aircraft} dataset \cite{airbus-aircraft-detection_dataset} consists of 109 high-resolution images that capture airplanes at various airports around the world. The images are taken at airport gates or tarmacs and are categorized into two folders: `images' and `extras'. The `images' folder contains 103 pictures extracted from Pleiades imagery, offering a resolution of approximately 50 cm. Each image is stored as a JPEG file with dimensions of $2560 \times 2560$ pixels, corresponding to a ground area of 1280 m$^2$. In particular, the dataset includes snapshots of certain airports taken on different dates, allowing researchers to explore temporal variations. Some images in the dataset also exhibit challenging weather conditions such as fog or clouds. Additionally, the `extras' folder provides a separate set of images that can be used for testing purposes, ensuring the evaluation of algorithms on completely unseen data.

%Reference:
%\href{https://www.kaggle.com/datasets/airbusgeo/airbus-aircrafts-sample-dataset}{Link}

\subsection{HRPlanesv2 Dataset}

The Google Earth \textbf{HRPlanesv2} \cite{article_1107890} dataset is a comprehensive collection of high-resolution aerial images for aircraft detection research. It comprises 2120 images sourced from Google Earth, showcasing airports from diverse regions and serving different purposes, including civil, military, and joint airports. These images offer a rich variety of aircraft instances, providing an extensive dataset for training and evaluation. Each image is stored as a JPEG file with dimensions of $4800 \times 2703$ pixels, ensuring detailed representations of the airport scenes. To facilitate object detection tasks, the dataset includes precise labels for 14,335 aircraft instances, provided in the YOLO annotation format. Moreover, the dataset is divided into three subsets: 70\% for training, 20\% for validation, and 10\% for testing, enabling researchers to assess and compare the performance of their algorithms accurately.

%Reference:
%\href{https://eod-grss-ieee.com/dataset-detail/ak1BclhJbkpuUkh5Uitmd3B5L2hNQT09}{Link}

\subsection{RarePlanes Dataset}

The \textbf{RarePlanes} dataset \cite{DBLP:journals/corr/abs-2006-02963} comprises both real and synthetic satellite imagery. Developed by CosmiQ Works and AI.Reverie, this dataset aims to evaluate the efficacy of AI.Reverie's synthetic data in enhancing computer vision algorithms for aircraft detection in satellite imagery. The number of the real images in the dataset comprises 253 Maxar WorldView-3 satellite scenes, taken at 112 distinct locations and spanning an impressive 2142 km². These scenes contain hand-annotated aircraft instances, totaling 14,700 annotations. In addition, the dataset includes 50,000 synthetic satellite images generated using AI.Reverie's advanced simulation platform. These synthetic images feature approximately 630,000 aircraft annotations, providing a valuable resource to explore the benefits of synthetic data in overhead aircraft detection.

%Reference:
%\href{https://www.cosmiqworks.org/rareplanes-public-user-guide}{Link}
\subsection{GDIT Dataset}

The \textbf{GDIT} Aerial Airport dataset \cite{GDIT-airport_dataset} is a specialized collection of aerial images that focuses on parked airplanes at airports. It presents an opportunity for researchers to explore aircraft detection algorithms in the context of airport environments. The dataset consists of 338 high-quality images with a resolution of 600 $\times$ 600 pixels, which are further categorized into training, validation, and testing subsets with 236, 68, and 34 images, respectively. Notably, some of the training images exhibit variations such as different filters, zoom levels, or rotations, resulting in an expanded dataset of 810 images. The dataset offers a unified classification label for all types of airplanes, simplifying the detection task.

%Reference:
%\href{https://universe.roboflow.com/gdit/aerial-airport}{Link}

\subsection{Planesnet Dataset}

The \textbf{Planesnet} dataset \cite{robert_hammell_2023} provides an extensive collection of satellite imagery extracted from Planet satellites, focusing on multiple airports in California. This dataset comprises 32 000 RGB images, each measuring $20 \times 20$ pixels. The images are meticulously labeled as either 'plane' or 'no-plane' enabling researchers to train and evaluate aircraft detection algorithms. Derived from PlanetScope full-frame visual scene products, the dataset ensures an orthorectified 3 m pixel size, capturing fine-grained details. The Planesnet dataset is available in two formats: a zipped directory containing the PNG images and a JSON file containing corresponding metadata. Each image is accompanied by a filename that includes the label, scene ID, and metadata such longitude, and latitude coordinates.

%Reference:
%\href{https://www.kaggle.com/datasets/rhammell/planesnet}{Link}

\subsection{Flying Airplanes Dataset}

The dataset of \textbf{Flying airplanes on satellite images} \cite{3mbt-tb11-21} offers valuable resources for research related to the detection of aircraft in satellite imagery. It includes 180 satellite images covering areas of interest surrounding 30 European airports. The dataset incorporates ground-truth annotations of flying airplanes, which can be used to support various research investigations. These annotations serve as a reference for developing and evaluating algorithms for flying airplane detection. The dataset comprises modified Sentinel-2 data processed by Euro Data Cube, providing high-quality satellite imagery suitable for multiple applications.

%Reference:
%\href{https://ieee-dataport.org/open-access/dataset-detecting-flying-airplanes-satellite-images}{Link}

\subsection{OPT-Aircraft Dataset}

The \textbf{OPT-Aircraft V1.0} \cite{chen2020dataset} dataset focuses on the identification of aircraft groups in remote sensing images. It includes 3594 airplane images obtained from various public datasets, such as DIOR, UCA AOD, NWPU VHR-10, DOTA, and Google Earth. The dataset encompasses seven aircraft groups categorized based on wings and propellers, further divided into fourteen sub-groups considering aircraft color and engine position. The dataset, stored in PNG format, consists of 3594 files with a compressed size of 69.3MB.

%%%%%%%%%%%%%%%%%%%%%%%%%%%%%%
%%%%%%%%% SECTION 4 %%%%%%%%%%
%%%%%%%%%%%%%%%%%%%%%%%%%%%%%%
\section{Literature Review: Object Detection}\label{sec:OD_survey}
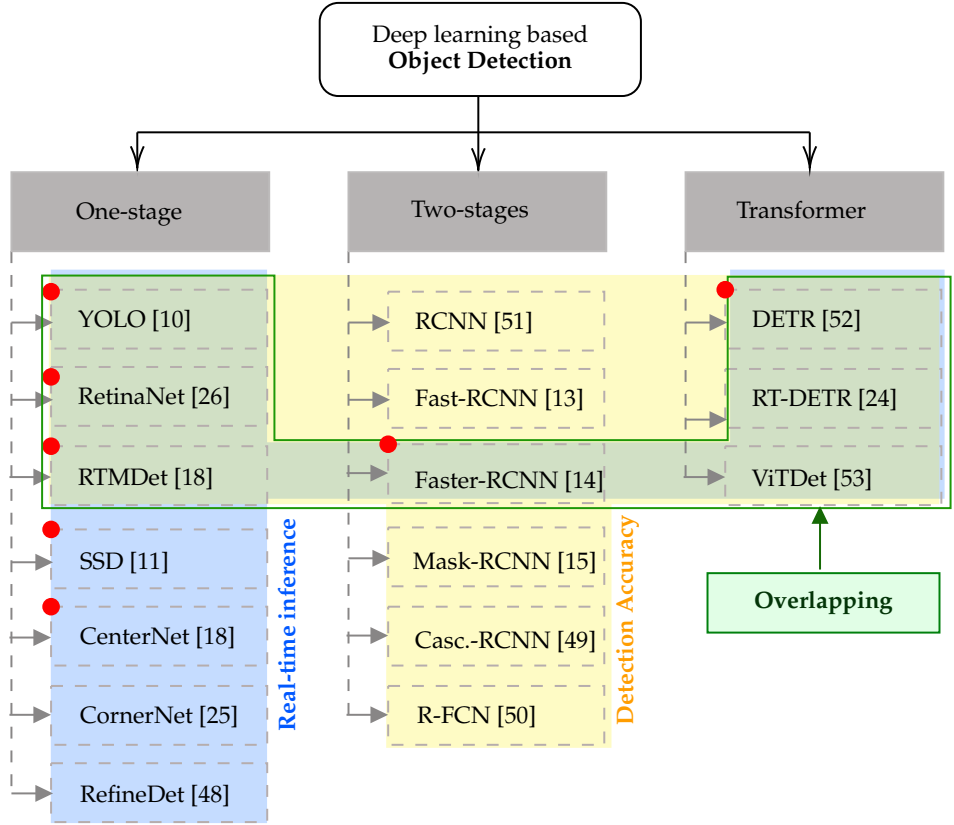
\begin{figure}[h!]
    \tikzset{every picture/.style={line width=0.75pt}} %set default line width to 0.75pt        
    
    \begin{tikzpicture}[x=0.75pt,y=0.75pt,yscale=-1,xscale=1]
    %uncomment if require: \path (0,494); %set diagram left start at 0, and has height of 494
    
    %Shape: Path Data [id:dp16530874237467064] 
    \draw  [draw opacity=0][fill={rgb, 255:red, 0; green, 105; blue, 255 }  ,fill opacity=0.23 ] (218.8,143.29) -- (218.8,230.44) -- (452.31,230.44) -- (452.31,143.29) -- (560.5,143.29) -- (560.5,258.95) -- (558.86,258.95) -- (558.86,298.14) -- (449.86,298.14) -- (449.86,258.95) -- (218.8,258.95) -- (218.8,422.67) -- (109.71,422.67) -- (109.71,143.29) -- (218.8,143.29) -- cycle ;
    %Shape: Path Data [id:dp5562799690435825] 
    \draw  [draw opacity=0][fill={rgb, 255:red, 254; green, 239; blue, 4 }  ,fill opacity=0.23 ] (110.71,143.45) -- (559.86,143.45) -- (559.86,333.14) -- (430,333.14) -- (430,259.14) -- (394.5,259.14) -- (394.5,382.12) -- (280.86,382.12) -- (280.86,259.14) -- (110.71,259.14) -- (110.71,143.45) -- cycle ;
    %Straight Lines [id:da24470604384086703] 
    \draw    (326.33,52) -- (326.33,88) ;
    \draw [shift={(326.33,90)}, rotate = 270] [color={rgb, 255:red, 0; green, 0; blue, 0 }  ][line width=0.75]    (10.93,-3.29) .. controls (6.95,-1.4) and (3.31,-0.3) .. (0,0) .. controls (3.31,0.3) and (6.95,1.4) .. (10.93,3.29)   ;
    %Straight Lines [id:da09899180036274147] 
    \draw    (154.33,71) -- (154.33,87) ;
    \draw [shift={(154.33,89)}, rotate = 270] [color={rgb, 255:red, 0; green, 0; blue, 0 }  ][line width=0.75]    (10.93,-3.29) .. controls (6.95,-1.4) and (3.31,-0.3) .. (0,0) .. controls (3.31,0.3) and (6.95,1.4) .. (10.93,3.29)   ;
    %Straight Lines [id:da43696553677231753] 
    \draw    (154.33,71) -- (493.5,71) ;
    %Shape: Rectangle [id:dp38836418049368415] 
    \draw  [color={rgb, 255:red, 182; green, 173; blue, 173 }  ,draw opacity=1 ][dash pattern={on 4.5pt off 4.5pt}] (110.86,150.43) -- (219.86,150.43) -- (219.86,180.14) -- (110.86,180.14) -- cycle ;
    %Shape: Rectangle [id:dp11593524639154884] 
    \draw  [color={rgb, 255:red, 182; green, 173; blue, 173 }  ,draw opacity=1 ][dash pattern={on 4.5pt off 4.5pt}] (110.86,189.43) -- (219.86,189.43) -- (219.86,219.14) -- (110.86,219.14) -- cycle ;
    %Shape: Rectangle [id:dp9778928181934494] 
    \draw  [color={rgb, 255:red, 182; green, 173; blue, 173 }  ,draw opacity=1 ][dash pattern={on 4.5pt off 4.5pt}] (110.86,229.43) -- (219.86,229.43) -- (219.86,259.14) -- (110.86,259.14) -- cycle ;
    %Shape: Rectangle [id:dp9187153637364849] 
    \draw  [color={rgb, 255:red, 182; green, 173; blue, 173 }  ,draw opacity=1 ][dash pattern={on 4.5pt off 4.5pt}] (110.86,271.43) -- (219.86,271.43) -- (219.86,301.14) -- (110.86,301.14) -- cycle ;
    %Shape: Rectangle [id:dp5624736093915077] 
    \draw  [color={rgb, 255:red, 182; green, 173; blue, 173 }  ,draw opacity=1 ][dash pattern={on 4.5pt off 4.5pt}] (110.86,310.43) -- (219.86,310.43) -- (219.86,340.14) -- (110.86,340.14) -- cycle ;
    %Shape: Rectangle [id:dp9207473469296052] 
    \draw  [color={rgb, 255:red, 182; green, 173; blue, 173 }  ,draw opacity=1 ][dash pattern={on 4.5pt off 4.5pt}] (110.86,350.43) -- (219.86,350.43) -- (219.86,380.14) -- (110.86,380.14) -- cycle ;
    %Shape: Rectangle [id:dp14501085939810365] 
    \draw  [color={rgb, 255:red, 182; green, 173; blue, 173 }  ,draw opacity=1 ][dash pattern={on 4.5pt off 4.5pt}] (110.86,389.43) -- (219.86,389.43) -- (219.86,419.14) -- (110.86,419.14) -- cycle ;
    %Straight Lines [id:da081868034955487] 
    \draw [color={rgb, 255:red, 136; green, 134; blue, 134 }  ,draw opacity=1 ] [dash pattern={on 4.5pt off 4.5pt}]  (90.86,131) -- (90.86,406.67) ;
    %Straight Lines [id:da7677246305546894] 
    \draw [color={rgb, 255:red, 136; green, 134; blue, 134 }  ,draw opacity=1 ]   (89.86,365) -- (107.86,365) ;
    \draw [shift={(110.86,365)}, rotate = 180] [fill={rgb, 255:red, 136; green, 134; blue, 134 }  ,fill opacity=1 ][line width=0.08]  [draw opacity=0] (8.93,-4.29) -- (0,0) -- (8.93,4.29) -- cycle    ;
    %Straight Lines [id:da7712283338522139] 
    \draw [color={rgb, 255:red, 136; green, 134; blue, 134 }  ,draw opacity=1 ]   (89.86,326) -- (107.86,326) ;
    \draw [shift={(110.86,326)}, rotate = 180] [fill={rgb, 255:red, 136; green, 134; blue, 134 }  ,fill opacity=1 ][line width=0.08]  [draw opacity=0] (8.93,-4.29) -- (0,0) -- (8.93,4.29) -- cycle    ;
    %Straight Lines [id:da8773996687885033] 
    \draw [color={rgb, 255:red, 136; green, 134; blue, 134 }  ,draw opacity=1 ]   (90.86,288) -- (108.86,288) ;
    \draw [shift={(111.86,288)}, rotate = 180] [fill={rgb, 255:red, 136; green, 134; blue, 134 }  ,fill opacity=1 ][line width=0.08]  [draw opacity=0] (8.93,-4.29) -- (0,0) -- (8.93,4.29) -- cycle    ;
    %Straight Lines [id:da09187879528183762] 
    \draw [color={rgb, 255:red, 136; green, 134; blue, 134 }  ,draw opacity=1 ]   (89.86,245) -- (107.86,245) ;
    \draw [shift={(110.86,245)}, rotate = 180] [fill={rgb, 255:red, 136; green, 134; blue, 134 }  ,fill opacity=1 ][line width=0.08]  [draw opacity=0] (8.93,-4.29) -- (0,0) -- (8.93,4.29) -- cycle    ;
    %Straight Lines [id:da21284601157360084] 
    \draw [color={rgb, 255:red, 136; green, 134; blue, 134 }  ,draw opacity=1 ]   (90.86,206) -- (108.86,206) ;
    \draw [shift={(111.86,206)}, rotate = 180] [fill={rgb, 255:red, 136; green, 134; blue, 134 }  ,fill opacity=1 ][line width=0.08]  [draw opacity=0] (8.93,-4.29) -- (0,0) -- (8.93,4.29) -- cycle    ;
    %Straight Lines [id:da1811671329644826] 
    \draw [color={rgb, 255:red, 136; green, 134; blue, 134 }  ,draw opacity=1 ]   (90.86,167) -- (108.86,167) ;
    \draw [shift={(111.86,167)}, rotate = 180] [fill={rgb, 255:red, 136; green, 134; blue, 134 }  ,fill opacity=1 ][line width=0.08]  [draw opacity=0] (8.93,-4.29) -- (0,0) -- (8.93,4.29) -- cycle    ;
    %Shape: Rectangle [id:dp6433738599667973] 
    \draw  [color={rgb, 255:red, 182; green, 173; blue, 173 }  ,draw opacity=1 ][dash pattern={on 4.5pt off 4.5pt}] (280.86,151.43) -- (389.86,151.43) -- (389.86,181.14) -- (280.86,181.14) -- cycle ;
    %Shape: Rectangle [id:dp2663445276060994] 
    \draw  [color={rgb, 255:red, 182; green, 173; blue, 173 }  ,draw opacity=1 ][dash pattern={on 4.5pt off 4.5pt}] (280.86,190.43) -- (389.86,190.43) -- (389.86,220.14) -- (280.86,220.14) -- cycle ;
    %Shape: Rectangle [id:dp7609232823900776] 
    \draw  [color={rgb, 255:red, 182; green, 173; blue, 173 }  ,draw opacity=1 ][dash pattern={on 4.5pt off 4.5pt}] (280.86,228.43) -- (389.86,228.43) -- (389.86,258.14) -- (280.86,258.14) -- cycle ;
    %Shape: Rectangle [id:dp18072422251734088] 
    \draw  [color={rgb, 255:red, 182; green, 173; blue, 173 }  ,draw opacity=1 ][dash pattern={on 4.5pt off 4.5pt}] (280.86,270.43) -- (389.86,270.43) -- (389.86,300.14) -- (280.86,300.14) -- cycle ;
    %Shape: Rectangle [id:dp13517890035891433] 
    \draw  [color={rgb, 255:red, 182; green, 173; blue, 173 }  ,draw opacity=1 ][dash pattern={on 4.5pt off 4.5pt}] (280.86,310.43) -- (389.86,310.43) -- (389.86,340.14) -- (280.86,340.14) -- cycle ;
    %Shape: Rectangle [id:dp6690777128393384] 
    \draw  [color={rgb, 255:red, 182; green, 173; blue, 173 }  ,draw opacity=1 ][dash pattern={on 4.5pt off 4.5pt}] (280.86,350.43) -- (389.86,350.43) -- (389.86,380.14) -- (280.86,380.14) -- cycle ;
    %Straight Lines [id:da7500828848398164] 
    \draw [color={rgb, 255:red, 136; green, 134; blue, 134 }  ,draw opacity=1 ] [dash pattern={on 4.5pt off 4.5pt}]  (260.86,131) -- (260.86,365) ;
    %Straight Lines [id:da3563282715433769] 
    \draw [color={rgb, 255:red, 136; green, 134; blue, 134 }  ,draw opacity=1 ]   (259.86,365) -- (277.86,365) ;
    \draw [shift={(280.86,365)}, rotate = 180] [fill={rgb, 255:red, 136; green, 134; blue, 134 }  ,fill opacity=1 ][line width=0.08]  [draw opacity=0] (8.93,-4.29) -- (0,0) -- (8.93,4.29) -- cycle    ;
    %Straight Lines [id:da32175943692644826] 
    \draw [color={rgb, 255:red, 136; green, 134; blue, 134 }  ,draw opacity=1 ]   (259.86,325) -- (277.86,325) ;
    \draw [shift={(280.86,325)}, rotate = 180] [fill={rgb, 255:red, 136; green, 134; blue, 134 }  ,fill opacity=1 ][line width=0.08]  [draw opacity=0] (8.93,-4.29) -- (0,0) -- (8.93,4.29) -- cycle    ;
    %Straight Lines [id:da2666250826569707] 
    \draw [color={rgb, 255:red, 136; green, 134; blue, 134 }  ,draw opacity=1 ]   (260.86,286) -- (278.86,286) ;
    \draw [shift={(281.86,286)}, rotate = 180] [fill={rgb, 255:red, 136; green, 134; blue, 134 }  ,fill opacity=1 ][line width=0.08]  [draw opacity=0] (8.93,-4.29) -- (0,0) -- (8.93,4.29) -- cycle    ;
    %Straight Lines [id:da8809440096863026] 
    \draw [color={rgb, 255:red, 136; green, 134; blue, 134 }  ,draw opacity=1 ]   (259.86,243) -- (277.86,243) ;
    \draw [shift={(280.86,243)}, rotate = 180] [fill={rgb, 255:red, 136; green, 134; blue, 134 }  ,fill opacity=1 ][line width=0.08]  [draw opacity=0] (8.93,-4.29) -- (0,0) -- (8.93,4.29) -- cycle    ;
    %Straight Lines [id:da19367749444889526] 
    \draw [color={rgb, 255:red, 136; green, 134; blue, 134 }  ,draw opacity=1 ]   (260.86,206) -- (278.86,206) ;
    \draw [shift={(281.86,206)}, rotate = 180] [fill={rgb, 255:red, 136; green, 134; blue, 134 }  ,fill opacity=1 ][line width=0.08]  [draw opacity=0] (8.93,-4.29) -- (0,0) -- (8.93,4.29) -- cycle    ;
    %Straight Lines [id:da9451981940832788] 
    \draw [color={rgb, 255:red, 136; green, 134; blue, 134 }  ,draw opacity=1 ]   (260.86,167) -- (278.86,167) ;
    \draw [shift={(281.86,167)}, rotate = 180] [fill={rgb, 255:red, 136; green, 134; blue, 134 }  ,fill opacity=1 ][line width=0.08]  [draw opacity=0] (8.93,-4.29) -- (0,0) -- (8.93,4.29) -- cycle    ;
    %Shape: Rectangle [id:dp6575853587361122] 
    \draw  [color={rgb, 255:red, 189; green, 187; blue, 187 }  ,draw opacity=1 ][fill={rgb, 255:red, 176; green, 173; blue, 173 }  ,fill opacity=0.92 ] (260.86,91.14) -- (390.86,91.14) -- (390.86,131) -- (260.86,131) -- cycle ;
    %Shape: Rectangle [id:dp7187655230499823] 
    \draw  [color={rgb, 255:red, 182; green, 173; blue, 173 }  ,draw opacity=1 ][dash pattern={on 4.5pt off 4.5pt}] (450.86,150.43) -- (559.86,150.43) -- (559.86,180.14) -- (450.86,180.14) -- cycle ;
    %Shape: Rectangle [id:dp6151614272242367] 
    \draw  [color={rgb, 255:red, 182; green, 173; blue, 173 }  ,draw opacity=1 ][dash pattern={on 4.5pt off 4.5pt}] (450.86,190.43) -- (559.86,190.43) -- (559.86,220.14) -- (450.86,220.14) -- cycle ;
    %Shape: Rectangle [id:dp7537374373328098] 
    \draw  [color={rgb, 255:red, 182; green, 173; blue, 173 }  ,draw opacity=1 ][dash pattern={on 4.5pt off 4.5pt}] (450.86,229.43) -- (559.86,229.43) -- (559.86,259.14) -- (450.86,259.14) -- cycle ;
    %Straight Lines [id:da6995170035451306] 
    \draw [color={rgb, 255:red, 136; green, 134; blue, 134 }  ,draw opacity=1 ] [dash pattern={on 4.5pt off 4.5pt}]  (430.86,131) -- (430.86,319) ;
    %Straight Lines [id:da9232471733097394] 
    \draw [color={rgb, 255:red, 136; green, 134; blue, 134 }  ,draw opacity=1 ]   (429.86,245) -- (447.86,245) ;
    \draw [shift={(450.86,245)}, rotate = 180] [fill={rgb, 255:red, 136; green, 134; blue, 134 }  ,fill opacity=1 ][line width=0.08]  [draw opacity=0] (8.93,-4.29) -- (0,0) -- (8.93,4.29) -- cycle    ;
    %Straight Lines [id:da6498628822575976] 
    \draw [color={rgb, 255:red, 136; green, 134; blue, 134 }  ,draw opacity=1 ]   (430.86,216) -- (448.86,216) ;
    \draw [shift={(451.86,216)}, rotate = 180] [fill={rgb, 255:red, 136; green, 134; blue, 134 }  ,fill opacity=1 ][line width=0.08]  [draw opacity=0] (8.93,-4.29) -- (0,0) -- (8.93,4.29) -- cycle    ;
    %Straight Lines [id:da784311593881319] 
    \draw [color={rgb, 255:red, 136; green, 134; blue, 134 }  ,draw opacity=1 ]   (430.86,167) -- (448.86,167) ;
    \draw [shift={(451.86,167)}, rotate = 180] [fill={rgb, 255:red, 136; green, 134; blue, 134 }  ,fill opacity=1 ][line width=0.08]  [draw opacity=0] (8.93,-4.29) -- (0,0) -- (8.93,4.29) -- cycle    ;
    %Shape: Rectangle [id:dp6604084571948672] 
    \draw  [color={rgb, 255:red, 189; green, 187; blue, 187 }  ,draw opacity=1 ][fill={rgb, 255:red, 176; green, 173; blue, 173 }  ,fill opacity=0.92 ] (430.86,91.14) -- (560.86,91.14) -- (560.86,131) -- (430.86,131) -- cycle ;
    %Shape: Circle [id:dp06996568678318771] 
    \draw  [color={rgb, 255:red, 255; green, 1; blue, 1 }  ,draw opacity=1 ][fill={rgb, 255:red, 255; green, 0; blue, 0 }  ,fill opacity=1 ] (106.93,194.43) .. controls (106.93,192.26) and (108.69,190.5) .. (110.86,190.5) .. controls (113.03,190.5) and (114.79,192.26) .. (114.79,194.43) .. controls (114.79,196.6) and (113.03,198.36) .. (110.86,198.36) .. controls (108.69,198.36) and (106.93,196.6) .. (106.93,194.43) -- cycle ;
    %Shape: Circle [id:dp17089035023072507] 
    \draw  [color={rgb, 255:red, 255; green, 1; blue, 1 }  ,draw opacity=1 ][fill={rgb, 255:red, 255; green, 0; blue, 0 }  ,fill opacity=1 ] (106.93,229.43) .. controls (106.93,227.26) and (108.69,225.5) .. (110.86,225.5) .. controls (113.03,225.5) and (114.79,227.26) .. (114.79,229.43) .. controls (114.79,231.6) and (113.03,233.36) .. (110.86,233.36) .. controls (108.69,233.36) and (106.93,231.6) .. (106.93,229.43) -- cycle ;
    %Shape: Circle [id:dp6012445089010949] 
    \draw  [color={rgb, 255:red, 255; green, 1; blue, 1 }  ,draw opacity=1 ][fill={rgb, 255:red, 255; green, 0; blue, 0 }  ,fill opacity=1 ] (106.93,271.43) .. controls (106.93,269.26) and (108.69,267.5) .. (110.86,267.5) .. controls (113.03,267.5) and (114.79,269.26) .. (114.79,271.43) .. controls (114.79,273.6) and (113.03,275.36) .. (110.86,275.36) .. controls (108.69,275.36) and (106.93,273.6) .. (106.93,271.43) -- cycle ;
    %Shape: Circle [id:dp9931239549368325] 
    \draw  [color={rgb, 255:red, 255; green, 1; blue, 1 }  ,draw opacity=1 ][fill={rgb, 255:red, 255; green, 0; blue, 0 }  ,fill opacity=1 ] (106.93,151.43) .. controls (106.93,149.26) and (108.69,147.5) .. (110.86,147.5) .. controls (113.03,147.5) and (114.79,149.26) .. (114.79,151.43) .. controls (114.79,153.6) and (113.03,155.36) .. (110.86,155.36) .. controls (108.69,155.36) and (106.93,153.6) .. (106.93,151.43) -- cycle ;
    %Shape: Circle [id:dp5838226350325246] 
    \draw  [color={rgb, 255:red, 255; green, 1; blue, 1 }  ,draw opacity=1 ][fill={rgb, 255:red, 255; green, 0; blue, 0 }  ,fill opacity=1 ] (106.93,310.43) .. controls (106.93,308.26) and (108.69,306.5) .. (110.86,306.5) .. controls (113.03,306.5) and (114.79,308.26) .. (114.79,310.43) .. controls (114.79,312.6) and (113.03,314.36) .. (110.86,314.36) .. controls (108.69,314.36) and (106.93,312.6) .. (106.93,310.43) -- cycle ;
    %Shape: Rectangle [id:dp6151470161650727] 
    \draw  [color={rgb, 255:red, 189; green, 187; blue, 187 }  ,draw opacity=1 ][fill={rgb, 255:red, 176; green, 173; blue, 173 }  ,fill opacity=0.92 ] (90.86,91.14) -- (220.86,91.14) -- (220.86,131) -- (90.86,131) -- cycle ;
    %Straight Lines [id:da40325209163946285] 
    \draw    (493.5,71) -- (493.5,87) ;
    \draw [shift={(493.5,89)}, rotate = 270] [color={rgb, 255:red, 0; green, 0; blue, 0 }  ][line width=0.75]    (10.93,-3.29) .. controls (6.95,-1.4) and (3.31,-0.3) .. (0,0) .. controls (3.31,0.3) and (6.95,1.4) .. (10.93,3.29)   ;
    %Straight Lines [id:da949311882671996] 
    \draw [color={rgb, 255:red, 21; green, 105; blue, 5 }  ,draw opacity=1 ][fill={rgb, 255:red, 15; green, 69; blue, 7 }  ,fill opacity=1 ]   (544.67,355.5) -- (544.67,298.67) ;
    \draw [shift={(544.67,295.67)}, rotate = 90] [fill={rgb, 255:red, 21; green, 105; blue, 5 }  ,fill opacity=1 ][line width=0.08]  [draw opacity=0] (8.93,-4.29) -- (0,0) -- (8.93,4.29) -- cycle    ;
    %Rounded Rect [id:dp961314433123694] 
    \draw  [color={rgb, 255:red, 0; green, 0; blue, 0 }  ,draw opacity=1 ][fill={rgb, 255:red, 255; green, 255; blue, 255 }  ,fill opacity=1 ] (246.33,15.07) .. controls (246.33,9.88) and (250.54,5.67) .. (255.73,5.67) -- (398.93,5.67) .. controls (404.12,5.67) and (408.33,9.88) .. (408.33,15.07) -- (408.33,43.27) .. controls (408.33,48.46) and (404.12,52.67) .. (398.93,52.67) -- (255.73,52.67) .. controls (250.54,52.67) and (246.33,48.46) .. (246.33,43.27) -- cycle ;
    %Shape: Circle [id:dp2856972036859933] 
    \draw  [color={rgb, 255:red, 255; green, 1; blue, 1 }  ,draw opacity=1 ][fill={rgb, 255:red, 255; green, 0; blue, 0 }  ,fill opacity=1 ] (446.93,150.43) .. controls (446.93,148.26) and (448.69,146.5) .. (450.86,146.5) .. controls (453.03,146.5) and (454.79,148.26) .. (454.79,150.43) .. controls (454.79,152.6) and (453.03,154.36) .. (450.86,154.36) .. controls (448.69,154.36) and (446.93,152.6) .. (446.93,150.43) -- cycle ;
    %Straight Lines [id:da824244132590011] 
    \draw [color={rgb, 255:red, 136; green, 134; blue, 134 }  ,draw opacity=1 ]   (90.86,404.67) -- (108.86,404.67) ;
    \draw [shift={(111.86,404.67)}, rotate = 180] [fill={rgb, 255:red, 136; green, 134; blue, 134 }  ,fill opacity=1 ][line width=0.08]  [draw opacity=0] (8.93,-4.29) -- (0,0) -- (8.93,4.29) -- cycle    ;
    %Shape: Circle [id:dp5372175811604232] 
    \draw  [color={rgb, 255:red, 255; green, 1; blue, 1 }  ,draw opacity=1 ][fill={rgb, 255:red, 255; green, 0; blue, 0 }  ,fill opacity=1 ] (276.93,228.43) .. controls (276.93,226.26) and (278.69,224.5) .. (280.86,224.5) .. controls (283.03,224.5) and (284.79,226.26) .. (284.79,228.43) .. controls (284.79,230.6) and (283.03,232.36) .. (280.86,232.36) .. controls (278.69,232.36) and (276.93,230.6) .. (276.93,228.43) -- cycle ;
    %Shape: Rectangle [id:dp19468419816430882] 
    \draw  [color={rgb, 255:red, 182; green, 173; blue, 173 }  ,draw opacity=1 ][dash pattern={on 4.5pt off 4.5pt}] (450.86,303.43) -- (559.86,303.43) -- (559.86,333.14) -- (450.86,333.14) -- cycle ;
    %Straight Lines [id:da9359244181521733] 
    \draw [color={rgb, 255:red, 136; green, 134; blue, 134 }  ,draw opacity=1 ]   (429.86,319) -- (447.86,319) ;
    \draw [shift={(450.86,319)}, rotate = 180] [fill={rgb, 255:red, 136; green, 134; blue, 134 }  ,fill opacity=1 ][line width=0.08]  [draw opacity=0] (8.93,-4.29) -- (0,0) -- (8.93,4.29) -- cycle    ;
    %Shape: Rectangle [id:dp03575069334922931] 
    \draw  [color={rgb, 255:red, 182; green, 173; blue, 173 }  ,draw opacity=1 ][dash pattern={on 4.5pt off 4.5pt}] (450.86,265.43) -- (559.86,265.43) -- (559.86,295.14) -- (450.86,295.14) -- cycle ;
    %Straight Lines [id:da6722426675527622] 
    \draw [color={rgb, 255:red, 136; green, 134; blue, 134 }  ,draw opacity=1 ]   (429.86,281) -- (447.86,281) ;
    \draw [shift={(450.86,281)}, rotate = 180] [fill={rgb, 255:red, 136; green, 134; blue, 134 }  ,fill opacity=1 ][line width=0.08]  [draw opacity=0] (8.93,-4.29) -- (0,0) -- (8.93,4.29) -- cycle    ;
    %Shape: Circle [id:dp3035047623443978] 
    \draw  [color={rgb, 255:red, 255; green, 1; blue, 1 }  ,draw opacity=1 ][fill={rgb, 255:red, 255; green, 0; blue, 0 }  ,fill opacity=1 ] (446.93,265.43) .. controls (446.93,263.26) and (448.69,261.5) .. (450.86,261.5) .. controls (453.03,261.5) and (454.79,263.26) .. (454.79,265.43) .. controls (454.79,267.6) and (453.03,269.36) .. (450.86,269.36) .. controls (448.69,269.36) and (446.93,267.6) .. (446.93,265.43) -- cycle ;
    
    % Text Node
    \draw (122,105) node [anchor=north west][inner sep=0.75pt]  [font=\small] [align=left] {One-stage};
    % Text Node
    \draw (371.19,383.96) node [anchor=north east] [inner sep=0.75pt]  [font=\small,color={rgb, 255:red, 255; green, 157; blue, 0 }  ,opacity=1 ,rotate=-359.45] [align=left] {\textbf{Detection}\\\textbf{Accuracy}};
    % Text Node
    \draw (195.35,426.63) node [anchor=north east] [inner sep=0.75pt]  [font=\small,color={rgb, 255:red, 0; green, 88; blue, 255 }  ,opacity=1 ,rotate=-359.98] [align=left] {\textbf{Real-time}\\\textbf{inference}};
    % Text Node
    \draw (455,104) node [anchor=north west][inner sep=0.75pt]  [font=\small] [align=left] {Transformer};
    % Text Node
    \draw (291,104) node [anchor=north west][inner sep=0.75pt]  [font=\small] [align=left] {Two-stages};
    % Text Node
    \draw  [color={rgb, 255:red, 0; green, 128; blue, 13 }  ,draw opacity=1 ][fill={rgb, 255:red, 0; green, 254; blue, 46 }  ,fill opacity=0.12 ]  (442.42,354.75) -- (560.42,354.75) -- (560.42,417.75) -- (442.42,417.75) -- cycle  ;
    \draw (501.42,386.25) node  [font=\small,color={rgb, 255:red, 3; green, 53; blue, 8 }  ,opacity=1 ] [align=left] {\begin{minipage}[lt]{77.41pt}\setlength\topsep{0pt}
    \begin{center}
    \textbf{Overlap:}\\Real-time and accuracy
    \end{center}
    
    \end{minipage}};
    % Text Node
    \draw (327.33,29.17) node  [font=\small] [align=left] {\begin{minipage}[lt]{87.44pt}\setlength\topsep{0pt}
    \begin{center}
    Deep learning based\\\textbf{Object Detection}
    \end{center}
    
    \end{minipage}};
    % Text Node
    \draw (124,319) node [anchor=north west][inner sep=0.75pt]  [font=\footnotesize] [align=left] {RetinaNet};
    % Text Node
    \draw (124,359) node [anchor=north west][inner sep=0.75pt]  [font=\footnotesize] [align=left] {CornerNet};
    % Text Node
    \draw (123,159) node [anchor=north west][inner sep=0.75pt]  [font=\footnotesize] [align=left] {YOLO};
    % Text Node
    \draw (123,198) node [anchor=north west][inner sep=0.75pt]  [font=\footnotesize] [align=left] {CenterNet};
    % Text Node
    \draw (123,239) node [anchor=north west][inner sep=0.75pt]  [font=\footnotesize] [align=left] {RTMDet};
    % Text Node
    \draw (124,281) node [anchor=north west][inner sep=0.75pt]  [font=\footnotesize] [align=left] {SSD};
    % Text Node
    \draw (124,399) node [anchor=north west][inner sep=0.75pt]  [font=\footnotesize] [align=left] {RefineDet};
    % Text Node
    \draw (293,321) node [anchor=north west][inner sep=0.75pt]  [font=\footnotesize] [align=left] {Casc.-RCNN};
    % Text Node
    \draw (294,359) node [anchor=north west][inner sep=0.75pt]  [font=\footnotesize] [align=left] {R-FCN};
    % Text Node
    \draw (293,160) node [anchor=north west][inner sep=0.75pt]  [font=\footnotesize] [align=left] {RCNN};
    % Text Node
    \draw (293,199) node [anchor=north west][inner sep=0.75pt]  [font=\footnotesize] [align=left] {Fast-RCNN};
    % Text Node
    \draw (292.86,240.43) node [anchor=north west][inner sep=0.75pt]  [font=\footnotesize] [align=left] {Faster-RCNN};
    % Text Node
    \draw (292,281) node [anchor=north west][inner sep=0.75pt]  [font=\footnotesize] [align=left] {Mask-RCNN};
    % Text Node
    \draw (464,160) node [anchor=north west][inner sep=0.75pt]  [font=\footnotesize] [align=left] {DETR};
    % Text Node
    \draw (463,199) node [anchor=north west][inner sep=0.75pt]  [font=\footnotesize] [align=left] {RT-DETR};
    % Text Node
    \draw (462.86,239.43) node [anchor=north west][inner sep=0.75pt]  [font=\footnotesize] [align=left] {ViTDet};
    % Text Node
    \draw (462.86,275.43) node [anchor=north west][inner sep=0.75pt]  [font=\footnotesize] [align=left] {Grounding Dino};
    % Text Node
    \draw (462.86,313.43) node [anchor=north west][inner sep=0.75pt]  [font=\footnotesize] [align=left] {MViTv2};
    
    \end{tikzpicture}

    %%%%%%%%%%%%%%
    \caption{{Classification of object detection methods based on (1) their architecture (one-stage, two-stage, and Transformer network), and (2) detection accuracy (in yellow) and real-time detection (blue). The red dot highlights the models that are implemented, trained, and validated in this work. The green outline (indicated as `Overlapping' in the image) groups the models that usually perform well in both accuracy and inference time response.
}}
    \label{fig:classification}
\end{figure}

Many object detection models have been proposed in the past few decades, the majority of which are based on CNNs \cite{Zhao2024}. So far, researchers have classified these models into two main categories: one- and two-stage architectures \cite{8741359, MITTAL2020104046}. However, the recent rise of the self-attention mechanism \cite{vaswani2017attention} has led to a new category of network architectures based on Transformers. Figure \ref{fig:classification} groups some of the most used and known models along with two types of classifications of deep learning models: (1) categorized based on their network architecture (one-stage, two-stage, or Transformer), or (2) based on their performance in real time and detection accuracy highlighted in blue and yellow, respectively. The overlapping area drawn in green features object detection models with balanced performances in both real-time detection and accuracy.

One-stage models \cite{liu2020deep} have been mainly known for their real-time deployment performance, because they are usually not computationally extensive and a single pass through the network is sufficient to produce estimations of object bounding boxes. However, the main limitation in this category is the detection accuracy, which might not be enough for some applications that require a very high detection confidence. In the case of two-stage object detection models, an additional stage is introduced to generate generic object proposals which make the model more efficient in its detection operation \cite{xie2024oriented}. The purpose of this stage is to produce candidate bounding boxes, which may not be highly accurate, and effectively exclude background areas from further processing. Subsequently, the next stage of the model undertakes the more computationally intensive tasks of classifying objects and refining the bounding boxes generated by the previous stage. The third category is the Transformer-based architecture \cite{9498550}, which seems to be a good balance between the accuracy of real-time detection, as it has been tested and compared on common datasets. This category has been tested on common large datasets and it leverages the use of the self-attention mechanism to produce reliable and accurate bounding box estimations while being able to perform real-time detection. 

The choice of object detection architecture depends on the application and the nature of images to process, which involves a trade-off between speed and accuracy \cite{lu_mimicdet_2020}.
 One-stage models tend to offer faster processing speeds, but may exhibit lower accuracy compared to their two-stage counterparts. The advantages of one-stage algorithms can be summarized as follows \cite{jordan_object_detection_2021,10.4108/eai.9-6-2022.174181, 8825470}:
\begin{itemize}
  \item Simplicity and Efficiency: One-stage detectors have a simpler architecture compared to two-stage algorithms. They directly predict object locations and class probabilities without the need for an intermediate proposal generation step. This simplicity leads to computational efficiency, as one-stage detectors can process images faster than two-stage detectors.
  \item Real-time Performance: One-stage detectors are designed to achieve real-time or near-real-time performance, making them suitable for applications where fast inference is crucial. These algorithms are commonly used in scenarios that require quick responses, such as autonomous driving, video analysis, and robotics \cite{lohia2021bibliometric}.
  \item Higher Recall: One-stage detectors tend to have a higher recall rate compared to two-stage detectors \cite{redmon_you_2016, liu_ssd_2016}. They are capable of detecting a larger number of objects in an image, including small or occluded objects, because of their dense and dense-like prediction strategies. This higher recall can be advantageous in applications where comprehensive object detection is more important than achieving extremely precise bounding box localization.
  \item Training Simplicity: One-stage detectors have a simpler training pipeline compared to two-stage detectors. They typically use a single-shot training strategy, where object locations and class predictions are directly regressed from the network output. This simplifies the training process, reduces the number of hyperparameters to tune, and requires fewer computational resources for training. Consequently, one-stage detectors are easier to implement and experiment with, especially for researchers or practitioners new to object detection algorithms.
\end{itemize}

%%%%%%%%%%%%%%%%%%%
%%%%%%%%%%%%%%%%%%%

On the other hand, two-stage models generally achieve higher accuracy but sacrifice some speed due to the additional processing stage. The advantages of two-stage algorithms over the others can be summarized with the following points:
\begin{itemize}
  \item Sampling Efficiency: Two-stage detectors employ a sampling mechanism to select a sparse set of region proposals, effectively eliminating a significant portion of the negative proposals. In contrast, one-stage detectors take a different approach by directly considering all regions in the image, which sometimes introduce class imbalance \cite{DBLP:journals/corr/abs-2009-11528}.
  \item Feature extraction: Two-stage detectors can allocate a larger head network for proposal classification and regression, as they only process a small number of proposals. This allows for the extraction of richer features, contributing to improved performance.
  \item Recall: Two-stage detectors take advantage of the RoIAlign \cite{he_mask_2017, DBLP:journals/corr/abs-2109-03495} operation to extract high-quality features from each proposal, ensuring location consistency. In contrast, in one-stage detectors, different region proposals may share the same feature, leading to coarse and spatially implicit representations that can cause feature misalignment.
  \item Accuracy: Two-stage detectors refine the location of the objects twice, once in each stage. Consequently, the bounding boxes generated by these models exhibit better accuracy compared to one-stage methods but at the expense of real-time performance.
\end{itemize}

Consequently, the trade-off between one-stage, two-stage, and Transformer-based architectures necessitates careful consideration based on the specific requirements of the application at hand as well as the nature of images. In this context, special attention is directed towards the detection of airplanes within remote sensing images. The choice of algorithms is guided by their widespread usage, popularity in performance when tested on other types of images, and availability as open-source implementations. This will include YOLO, CenterNet, RTMDet, SSD, RetinaNet, Faster-RCNN, and DETR.

% \subsection{{Models backbones}}
% {
% EXPLANATION OF RECURRENT OBJECT DETECTION AND SEGMENTATION BACKBONES SUCH AS RESNET \cite{he2015deepresiduallearningimage}, VGG, Swin transformers \cite{liu2021swintransformerhierarchicalvision} ... etc
% }

\subsection{{Models Backbones}}

{Backbones are the foundational neural network architectures used for feature extraction in computer vision models \cite{ELHARROUSS2024100645}, significantly influencing their accuracy and computational efficiency. They form the core of object detection and classification pipelines by providing hierarchical feature representations that are then used to identify objects both location and class at various scales and complexities. One of the first widely adopted backbones is the Very Deep Convolutional Network (VGG) \cite{simonyan2015deepconvolutionalnetworkslargescale}, introduced by Simonyan and Zisserman, which uses deep convolutional layers to build robust feature maps. Despite its simplicity and effectiveness, VGG networks are computationally intensive, making them less suitable for real-time applications. To address this, ResNet \cite{he2015deepresiduallearningimage} was proposed by He et al., introducing residual connections to ease deep network training and improve performance. ResNet variants, such as ResNet-50 and ResNet-101, remain fundamental in many object detection frameworks such as Faster R-CNN \cite{ren_faster_2015} and RetinaNet \cite{RetinaNet_Lin} due to their balance of efficiency and representational power. The field advanced later with feature pyramid designs, as seen in the Feature Pyramid Network (FPN) \cite{lin2017featurepyramidnetworksobject} layered atop backbones like ResNet. FPN enhances multi-scale feature extraction, making it particularly valuable for detecting objects of varying sizes, as implemented in YOLOv5 \cite{diagnostics13132280}. High-Resolution Networks (HRNet) \cite{wang2020deephighresolutionrepresentationlearning} further improved upon this by maintaining high-resolution representations throughout the network, which proved advantageous for both semantic segmentation and object detection tasks. HRNet demonstrates high performance in models such as Mask R-CNN \cite{he_mask_2017} and Cascade R-CNN \cite{cai2018cascade}. 

Transformer-based architectures have emerged as another transformative innovation through Swin Transformers \cite{liu2021swintransformerhierarchicalvision}, which divide input images into patches and apply hierarchical attention mechanisms. These architectures are particularly effective in handling long-range dependencies and global context, making them prominent in modern frameworks like DETR \cite{lv2023detrs} and DINO \cite{zhang2022dinodetrimproveddenoising}. Swin Transformers also integrate well with dense prediction tasks, achieving state-of-the-art results across various benchmarks. Furthermore, YOLOv10 \cite{THU_MIGyolov10} represents one of the latest advances in the YOLO family for object detection, combining traditional convolutional approaches with Transformer-based modules to enhance attention mechanisms and reduce computational redundancy.}

\subsection{{One-Stage Models}}
\subsubsection{{You Only Look Once}}
The You Only Look Once (YOLO) framework \cite{redmon_you_2016} has emerged as a popular deep learning-based object detection algorithm that revolutionized real-time object detection tasks. It presents a unified approach to object detection by formulating it as a regression problem, allowing the model to predict bounding boxes and class probabilities of multiple objects in a single pass through the network.

\begin{figure}[H]
       
        \includegraphics[width=\textwidth,scale=1]{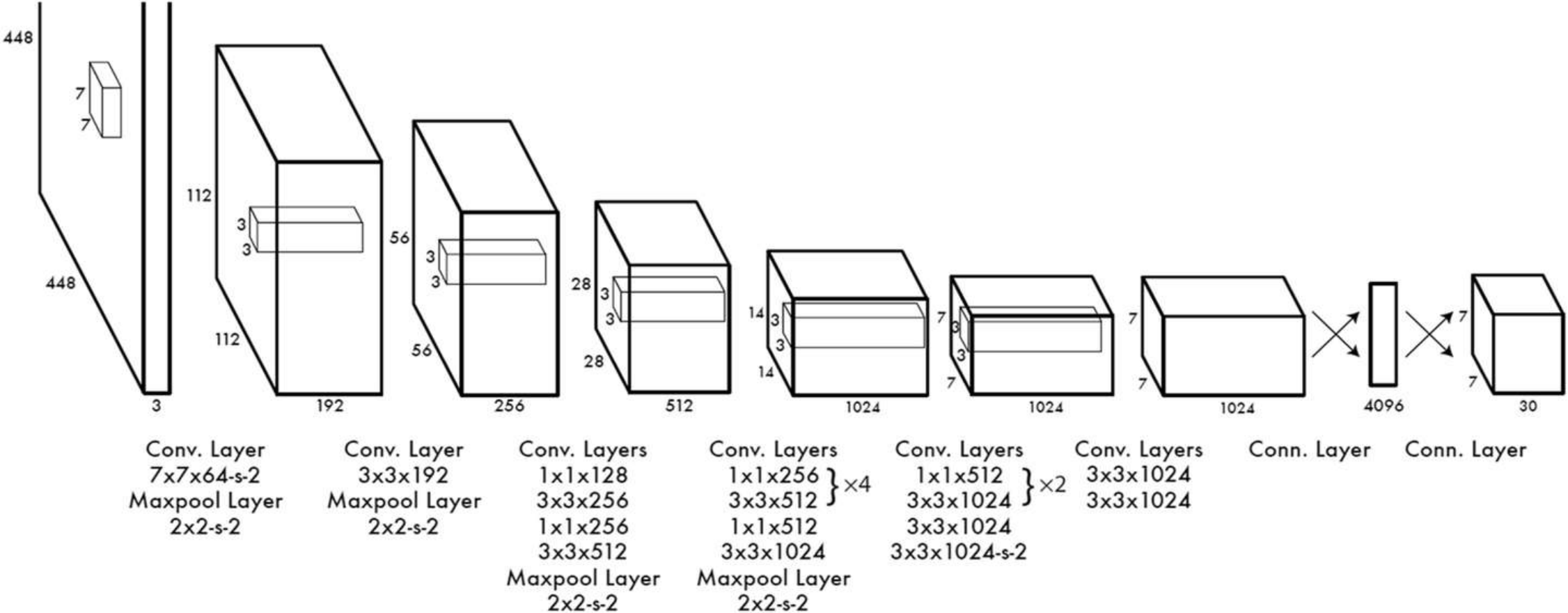}
        \caption{Basic YOLO architecture. Reproduced from \cite{diwan2023object}.}
        \label{fig:yoloarchitechture}
    \end{figure}

In the early stages, YOLOv2 \cite{8100173} introduced batch normalization and high-resolution classifiers to improve detection performance. YOLOv3 \cite{redmon2018yolov3} further refined the architecture by incorporating skip connections and multi-scale prediction, enhancing both accuracy and localization capabilities. The introduction of CSPDarknet, SPP, PAN, and Mish activation in YOLOv4 \cite{alexey2020yolov4} led to significant improvements in the network architecture. YOLOv5 \cite{2022zndo...7347926J}, based on scaled-YOLOv4 \cite{wang2021scaledyolov4}, introduced anchor-free object detection and a new architecture. This version expanded the options available by providing models of varying sizes, allowing users to balance speed and accuracy according to their specific requirements  (Figure \ref{fig:yoloarchitechture}).

{Some of the most known stable versions are YOLOv8 \cite{jocher2023yolo} and YOLO-NAS \cite{supergradientsyolonas}, which introduced the concept of neural architecture search (NAS) \cite{chitty2023nas} to automatically design network architectures and achieve state-of-the-art performance in object detection tasks. More recently, YOLOv10 \cite{THU_MIGyolov10} introduced significant innovations to improve both efficiency and accuracy in real-time object detection. A key change is the move away from traditional non-maximum suppression (NMS) in favor of a consistent dual-assignment strategy. This change enables a more efficient end-to-end deployment with lower inference latency. Furthermore, YOLOv10 \cite{THU_MIGyolov10} incorporates Transformer-based enhancements, allowing for more effective attention mechanisms. Such a holistic model is designed to optimize various components for both computational efficiency and detection performance, resulting in reduced redundancy. This optimization leads to a model with  46\% less latency, and 25\% fewer parameters than YOLOv8 \cite{jocher2023yolo} %EE: check meaning retained
	 while still achieving state-of-the-art results. The continuous evolution of the YOLO framework highlights the trade-offs between speed and accuracy, necessitating consideration of the specific application requirements when selecting an appropriate YOLO model.}

\subsubsection{{Single-Shot Detection}}

\begin{figure}[H]
      
        \includegraphics[width=\textwidth,scale=1.5]{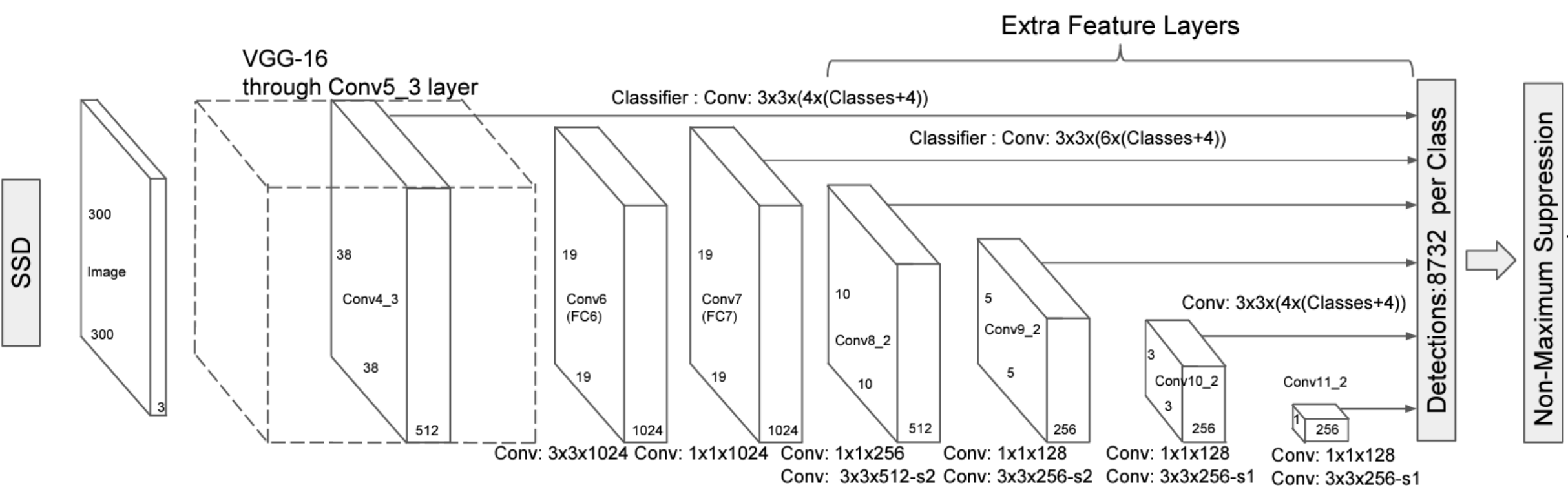}
        \caption{SSD architecture diagram. Reproduced from \cite{10.1007/978-3-319-46448-0_2}.}
        \label{fig:ssdarchitechture}
    \end{figure}

The Single-Shot Detector (SSD) \cite{10.1007/978-3-319-46448-0_2} is an efficient and accurate object detection algorithm that introduces a unified framework for single-pass detection. In theory, SSD addresses the challenge of detecting objects at various scales including small targets \cite{Ma2023} and aspect ratios, using predefined anchor boxes \cite{s21103569_anchor}. The SSD architecture consists of three main components: a base network, convolutional feature maps, and convolutional predictors (Figure \ref{fig:ssdarchitechture}). The base network, usually a pre-trained CNN \cite{9429420_pretrainedcnn}, acts as a feature extractor, generating high-level feature maps with different spatial resolutions. These feature maps are then processed by convolutional predictors, which predict the presence and location of objects for the anchor boxes. Each predictor corresponds to a specific feature map and produces class scores and bounding box offsets. The anchor boxes, distributed densely throughout the feature map, serve as reference boxes for object detection. To capture objects at multiple scales, SSD employs feature maps from different stages of the base network. Higher-resolution feature maps are effective at detecting small objects, while lower-resolution ones are suitable for larger objects. The predictions from each feature map are combined to generate final class predictions and refined bounding box coordinates. During training, SSD uses a multitask loss function \cite{kendall2018multitask} that optimizes the model considering both localization loss (Smooth L1 loss) and classification loss (softmax loss) \cite{terven2023loss}. The localization loss penalizes the discrepancy between predicted and ground truth bounding box coordinates, while the classification loss encourages accurate class predictions. The loss is computed for positive and negative samples, including hard negatives based on confidence scores.

Although SSD offers a good balance between accuracy and efficiency with its variable feature map resolutions and fixed anchor boxes, it does have limitations \cite{DBLP:journals/corr/JeongPK17}. One limitation is the dependence on predefined anchor boxes, which may not adequately cover objects with extreme aspect ratios or unconventional shapes. Scaling the anchor boxes can help address this, but it introduces additional computational overhead. Another limitation is the challenge of handling tiny objects \cite{kumar_object_2020}. Since SSD relies on a limited number of feature maps, it may struggle to accurately detect small objects due to the loss of fine-grained details at higher resolutions. This can result in reduced localization accuracy and increased false negatives for small or densely packed objects. Furthermore, SSD's fixed anchor boxes limit its ability to handle objects at arbitrary scales and aspect ratios.

\subsubsection{{RetinaNet Framework}}
\begin{figure}[H]
       
        \includegraphics[width=\textwidth,scale=1]{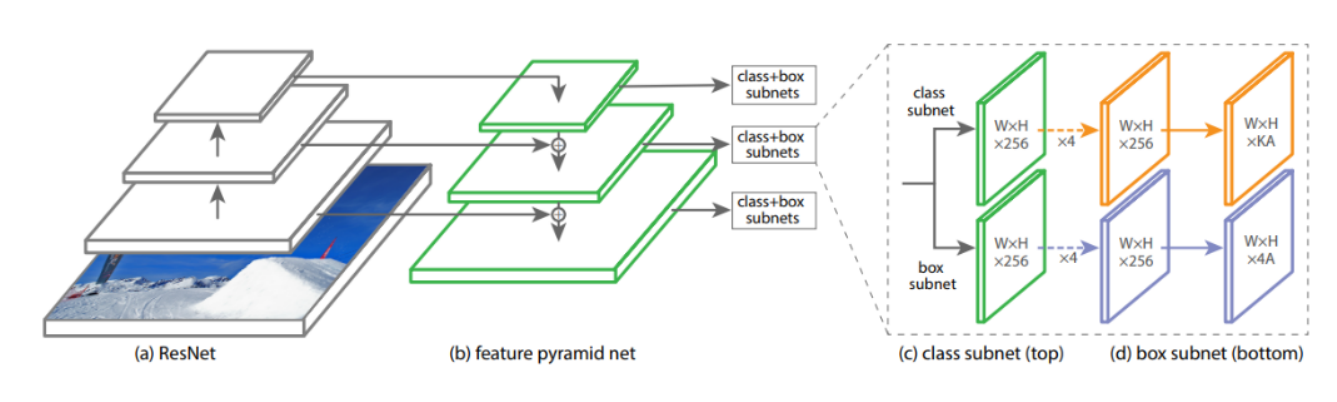}
        \caption{One-stage RetinaNet architecture. Reproduced from \cite{lin_focal_2020}. (\textbf{a}) ResNet \cite{he2016deep} backbone. (\textbf{b}) Generation of multi-scale convolutional pyramid. This is attached to two subnetworks: (\textbf{c}) anchor box classification and (\textbf{d}) anchor box regression to ground-truth bounding box.} 
        \label{fig:retinanetarchitechture}
    \end{figure}

RetinaNet is a one-stage object detection model that was introduced by Tsung-Yi Lin et al. \cite{RetinaNet_Lin} designed to address the extreme foreground--background class imbalance encountered during training of dense detectors \cite{Karol2024}. In object detection, the goal is to detect objects of interest in an image and localize them by drawing bounding boxes around them. However, the vast majority of the image is typically background, and there are usually many more negative examples (background) than positive examples (objects of interest). This class imbalance can make it difficult for the detector to learn to distinguish between objects and background, and can lead to poor performance. RetinaNet is a single unified network composed of a backbone network and two task-specific subnetworks (Figure \ref{fig:retinanetarchitechture}). The backbone is responsible for computing a convolutional feature map over an entire input image and is an off-the-shelf convolutional network. The first subnetwork is a dense prediction subnet that produces a fixed number of object detections of different scales and aspect ratios at each spatial location in a feature map. The second subnetwork is a set of class-specific subnets that further refine the predictions of the first subnet.

This model uses a novel focal loss function that down-weights the contribution of easy examples during training to focus on hard examples and prevents the vast number of easy background examples from overwhelming the detector. The focal loss is a dynamically scaled cross-entropy loss, where the scaling factor decays to zero as confidence in the correct class increases. This loss function is designed to address the mechanisms used by other detectors to address class imbalance, such as biased minibatch sampling and object proposal mechanisms, in a one-stage detection system directly via the loss function.

The obtained results showed good performance compared to previous one-stage and two-stage detectors, including the fastest reported R-CNN system, on the COCO dataset. It achieves state-of-the-art performance on both COCO detection tasks, with a better COCO test-dev average precision while running at 5 fps. This is a significant improvement over previous state-of-the-art techniques for training one-stage detectors, such as training with sampling heuristics or hard example mining. RetinaNet is also designed to be efficient and scalable. It uses an efficient in-network feature pyramid that allows it to detect objects at multiple scales and resolutions, and it uses anchor boxes to improve localization accuracy. The anchor boxes are predefined boxes of different scales and aspect ratios that are placed at each spatial location in the feature map. The dense prediction subnet predicts the offsets and scales of the anchor boxes to generate object detections.

RetinaNet has been widely adopted in industry and academia and selected for a variety of applications, including object detection in autonomous driving, face detection, and medical image analysis. It has also inspired further research in the field of object detection, including the development of other novel loss functions and architectures.

\subsubsection{{CenterNet Framework}}
\begin{figure}[H]
     
        \includegraphics[width=\textwidth,scale=1]{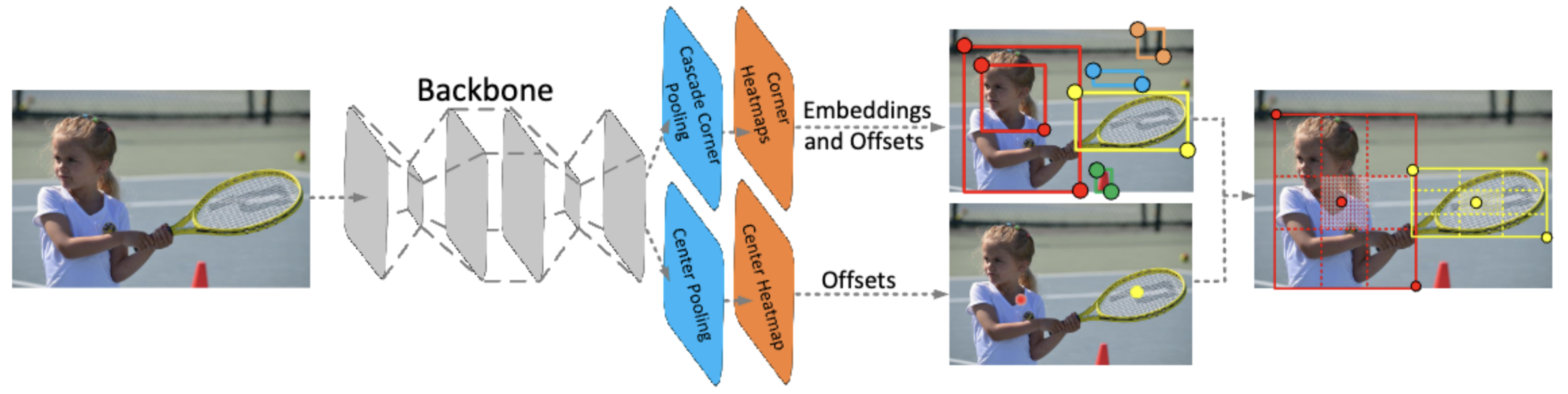}
        \caption{One-stage CenterNet architecture. Reproduced from \cite{duan_centernet_2019}. %MDPI: Please confirm whether the boxes of different colors require explanation.
}
        \label{fig:centernetarchitechture}
    \end{figure}

CenterNet \cite{duan_centernet_2019} is another real-time object detection algorithm designed to operate in real time, with an average inference time of 270 ms using a 52-layer hourglass backbone and 340 ms using a 104-layer hourglass backbone per image according to the author (Figure \ref{fig:centernetarchitechture}). CenterNet is inspired by the architecture of CornerNet \cite{DBLP:CornerNet_journals/corr/abs-1808-01244}, which is based on a one-stage keypoint-based detector, while introducing several novel components and strategies to enhance its effectiveness.

Unlike CornerNet, which detects object bounding boxes using pairs of keypoints, CenterNet introduces the concept of detecting each object as a triplet of keypoints. This innovation allows CenterNet to capture more comprehensive information about the objects, leading to improved detection performance while being fast for live inferencing. CenterNet incorporates two customized modules named cascade corner pooling and center pooling. These modules play crucial roles in enriching information collected by both top-left and bottom-right corners and providing more recognizable information at the central regions, respectively. These modules contribute to the improved performance of CenterNet by enhancing the detection of object keypoints and bounding boxes. In their benchmark study \cite{duan_centernet_2019}, the authors stated that this proposed architecture achieves significant improvements over existing one-stage detectors, typically an average precision of 47.0\% on the MS-COCO dataset, outperforming the previously proposed one-stage detectors by at least 4.9\%. Additionally, it demonstrates comparable performance to the two-stage detectors while maintaining a faster inference speed thanks to the effective reduction of incorrect bounding boxes, particularly for small objects. It achieves notable improvements in the detection of small objects, with an average precision (AP) improvements of 5.5\% to 8.1\% for different backbone configurations. This reduction in incorrect bounding boxes is attributed to the effectiveness of CenterNet in modeling center information using center keypoints. 

\subsubsection{RTMDet}
\begin{figure}[H]
        \centering
        \includegraphics[width=\textwidth,scale=1]{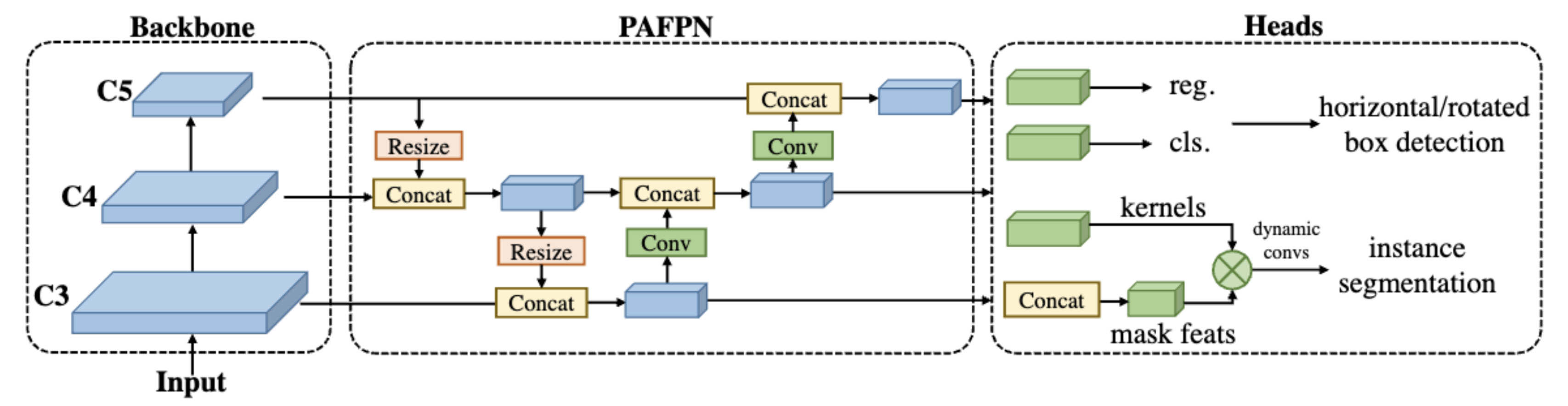}
        \caption{One-stage RTMDet architecture. Reproduced from \cite{lyu2022RTMDet}.
}
        \label{fig:RTMDetarchitechture}
    \end{figure}

RTMDet \cite{lyu2022RTMDet} is a groundbreaking real-time object detection model designed to achieve optimal efficiency without compromising accuracy. It belongs to the family of fully convolutional single-stage detectors \cite{chen2019mmdetection} such as the YOLO series. RTMDet \cite{lyu2022RTMDet} operates as a one-stage detector, allowing it to quickly recognize and locate objects in real-world scenarios such as autonomous driving, robotics, and drones (Figure \ref{fig:RTMDetarchitechture}). The model is engineered to push the boundaries of the YOLO series by introducing a new family of real-time models for object detection. Notably, RTMDet is capable of performing instance segmentation and rotated object detection, features that were previously unexplored.

RTMDet operates on a model architecture that emphasizes efficiency and compatibility in both the backbone and neck. The basic building block of the model consists of large-kernel depth-wise convolutions \cite{ding2022scaling}, which contribute to its ability to capture global context effectively. This architectural choice enhances the model's capacity to recognize and localize objects with high precision. Furthermore, RTMDet incorporates soft labels during the calculation of matching costs in the dynamic label assignment process \cite{dynamic_label_Zhang_2022}, leading to an improved accuracy. The combination of these architectural features and training techniques culminates in an object detector that achieves exceptional performance. RTMDet's macro architecture follows the one-stage object detector paradigm, and balances the depth, width, and resolution of the model to optimize efficiency. Additionally, the model is designed to be versatile, allowing for easy extension to instance segmentation and rotated object detection tasks with minimal modifications.

In comparison to the state-of-the-art industrial detectors, the authors of \cite{lyu2022RTMDet} demonstrate the remarkable performance of RTMDet in terms of both speed and accuracy. The model achieves an impressive 52.8\% AP on the COCO dataset while operating at over 300 frames per second (FPS) on an NVIDIA 3090 GPU. This outperforms the current mainstream industrial detectors, showcasing the superior parameter--accuracy trade-off of RTMDet. %EE: check meaning retained
The model's versatility is evident in its ability to deliver optimal performance across various application scenarios, offering different model sizes for different object recognition tasks.

\subsection{{Two-Stage Models}}
\subsubsection*{{Region-Based CNNs}}

\begin{figure}[H]
      \centering
        \includegraphics[width=120mm,scale=1]{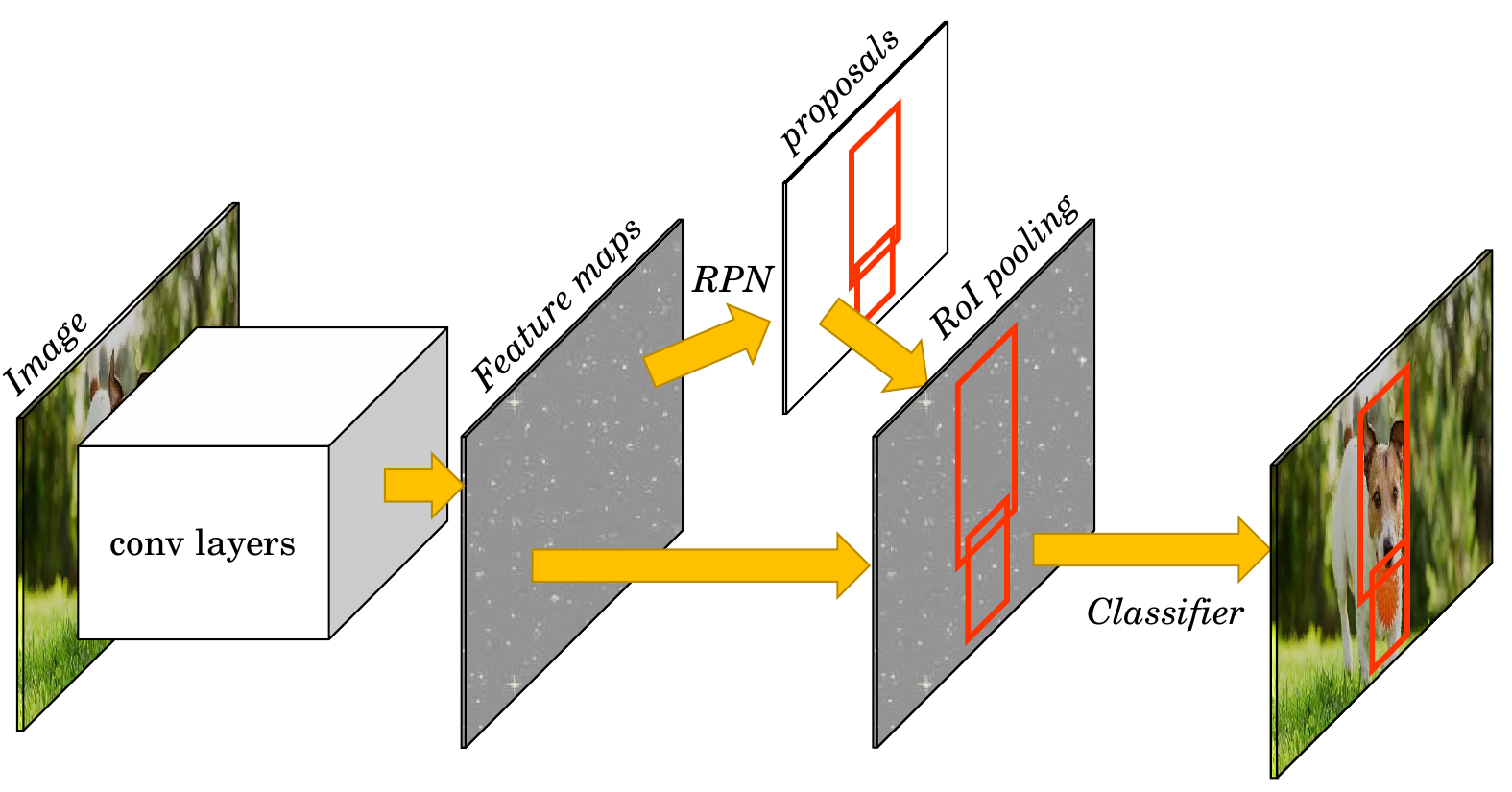}
        \caption{Faster RCNN architecture. Reproduced from \cite{ren_faster_2015}.}
        \label{fig:frcnnarchitechture}
    \end{figure}

The introduction of the Region-based Convolutional Neural Network (R-CNN) \cite{girshick_rich_2014} marked a significant milestone in the development of object detection techniques \cite{Archana2024}, showcasing the substantial improvements that Convolutional Neural Networks (CNNs) can bring to detection performance. R-CNN introduced the concept of utilizing CNNs \cite{yamashita_2018_cnn_overview} in combination with a class-agnostic region proposal \cite{Jaiswal2020ClassagnosticOD} module to transform object detection into a classification and localization problem. The detection process starts with a mean-subtracted input image, which is fed through the region proposal (RPN) module. This module employs techniques such as Selective Search \cite{uijlings_2013_selectivesearch} to identify regions within the image that have a higher likelihood of containing objects. Approximately 2000 object candidates are generated on the basis of this region proposal step. These candidates are then warped and passed through a CNN network, such as the widely used ImageNet \cite{krizhevsky_imagenet_2012}, to extract a 4096-dimensional feature vector for each proposal. The feature vectors obtained from the CNN are then used as inputs for class-specific support vector machines (SVMs) \cite{svm_708428}, which have been trained beforehand. The SVMs generate confidence scores for each candidate region, aiding in the classification process. To refine the results, non-maximum suppression (NMS) is applied based on the intersection over union (IoU) and class information. Once the class has been identified, a trained bounding box regressor is employed to predict the precise bounding box coordinates, including the center coordinates, width, and height of the object (Figure \ref{fig:frcnnarchitechture}).

However, despite its groundbreaking contributions, R-CNN has several limitations \cite{zou2023object, article_limitation_rs_algo}. The R-CNN training process is complex and multistage. It involves pre-training the CNN with a large classification dataset, followed by fine-tuning on domain-specific images that undergo mean subtraction and warping to align with the proposals. The CNN classification layer is replaced with a randomly initialized N + 1-way classifier, where N represents the number of classes, and stochastic gradient descent (SGD) is utilized for optimization. Additionally, separate SVMs and bounding box regressors need to be trained for each class, adding to the computational complexity.

Although R-CNN is capable of performing highly accurate object detection research, it suffered from slow inference times, taking approximately 47 s per image, and was resource-intensive in terms of both time and space \cite{girshick_rich_2014}. Training R-CNN models on small datasets took days to complete, even with shared computations. These limitations sparked the need for further advancements in object detection algorithms that could address these challenges and improve overall efficiency.

\subsection{{Transformer-Based Architectures}}

\subsubsection{{DETR: End-to-End Transformer}}

The advent of Transformer architectures has revolutionized the field of artificial intelligence by introducing a novel approach to processing sequential data. Unlike older neural network architectures, Transformers rely on self-attention mechanisms \cite{vaswani2017attention} to capture dependencies between input tokens, enabling them to effectively model long-range dependencies and capture complex patterns in the data. This ability to process sequences in parallel, rather than sequentially, has significantly improved the efficiency and effectiveness, especially of natural language processing (NLP) tasks \cite{LIN2022111}. The success of Transformers can be attributed to their ability to capture global context and relationships within the input data, making them particularly well suited for tasks that require an understanding of complex interdependencies.

Later on, researchers translated the use of Transformers in NLP to computer vision, including object detection operation. One pioneering algorithm is DETR (DEtection TRansformer) \cite{carion2020endtoend}, which belongs to the category of end-to-end object detection systems based on Transformers and bipartite matching loss for direct set prediction. Unlike traditional object detection methods, DETR does not fall into the conventional one-stage or two-stage detector categories. Instead, it introduces a novel approach by predicting all objects at once using a bipartite matching loss function, which uniquely assigns a prediction to a ground truth object and is invariant to a permutation of predicted objects. This unique approach simplifies the detection pipeline by eliminating the need for hand-designed components such as spatial anchors or non-maximal suppression \cite{zhao2019object}, making it optimal for both accuracy and real-time processing. DETR achieves this by leveraging Transformers with parallel decoding, as opposed to autoregressive decoding with recurrent neural networks, which was the focus of previous work.

\begin{figure}[h!]
      \centering
        \includegraphics[width=\textwidth,scale=1]{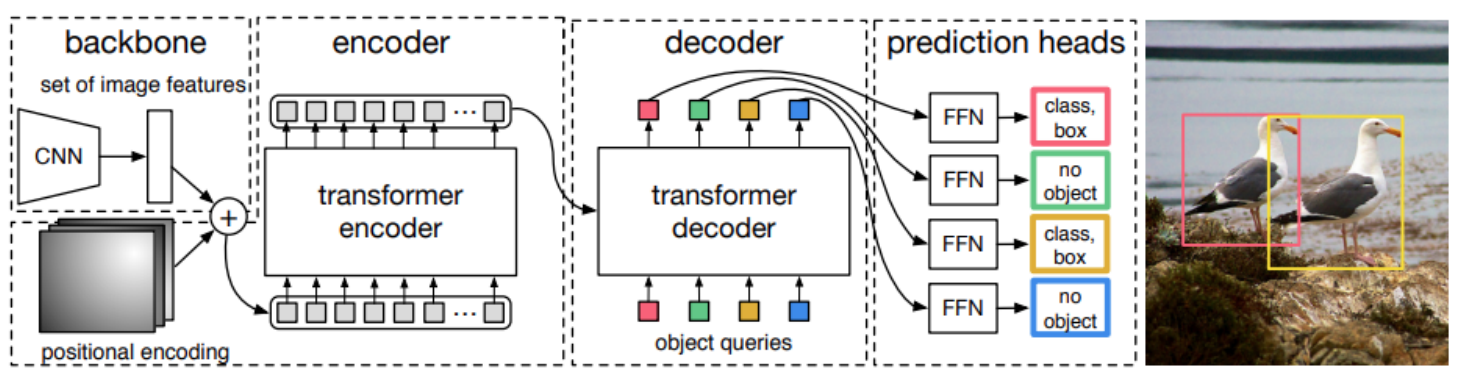}
        \caption{DETR architecture. Reproduced from \cite{carion2020endtoend}.}
        \label{fig:detrarchitechture}
    \end{figure}

The architecture of DETR is simple yet highly effective. It comprises three main components: a CNN backbone for feature extraction \cite{elharrouss2022backbonesreview}, an encoder--decoder Transformer %EE: check meaning retained
\cite{LIN2022111} for modeling relationships between feature representations of different detections, and a simple feed-forward network for making the final detection predictions (Figure \ref{fig:detrarchitechture}). The process begins with the extraction of a lower-resolution activation map from the input image using a conventional CNN backbone. This activation map is then passed through the Transformer encoder, along with spatial positional encodings that are added to queries and keys at every multi-head self-attention layer. The decoder receives queries, output positional encodings (object queries), and encoder memory, and produces the final set of predicted class labels and bounding boxes through multiple multi-head self-attention and decoder--encoder attention (FFNs). The simplicity and modularity of the DETR architecture make it easily implementable in any deep learning framework that provides a common CNN backbone and a Transformer architecture implementation.

In comparison to existing models, DETR has demonstrated remarkable performance on the challenging COCO object detection dataset. It achieves competitive results with the same number of parameters as Faster R-CNN, a widely used object detection model, achieving 42 AP on the COCO validation subset. Notably, DETR outperforms Faster R-CNN in terms of AP improvement, particularly in the context of direct set prediction. However, it lags behind in terms of small object AP. Additionally, DETR with a ResNet-101 \cite{app132413111} backbone shows comparable results to Faster R-CNN. The success of DETR can be attributed to its unique combination of bipartite matching loss \cite{GU2022104401} and Transformers with parallel decoding, which enables it to effectively model relations between feature representations of different detections and achieve competitive performance in object detection tasks.

\subsubsection{{Grounding DINO}}

\begin{figure}[h!]
    \centering
    \includegraphics[width=\textwidth,scale=1]{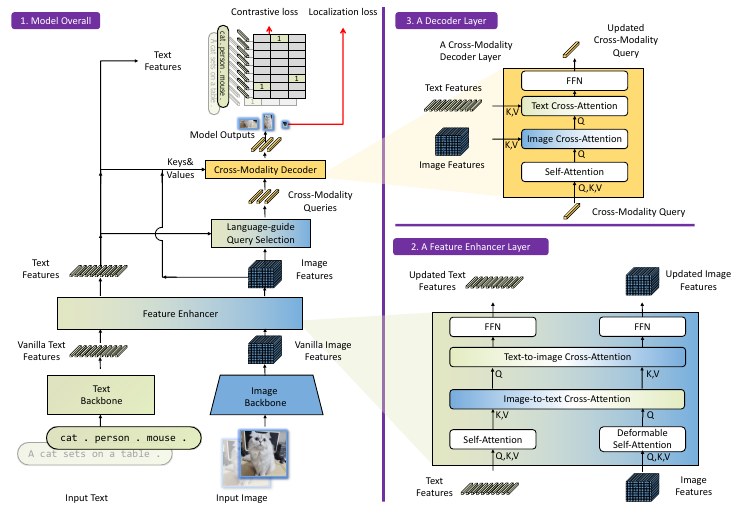}
    \caption{Grounding DINO architecture. Reproduced from \cite{liu2024groundingdinomarryingdino}.}
    \label{fig:groundingdinoarchitechture}
\end{figure}

Grounding DINO \cite{liu2024groundingdinomarryingdino} is a novel object detection model that merges the DINO (DEtection with INterpolation Optimization) architecture with grounded pre-training, aiming for effective open-set object detection. This integration allows the model to not only perform classical object detection but to identify arbitrary objects guided by textual descriptions or category names, expanding its utility in scenarios where predefined class categories are insufficient. The Grounding DINO model relies on a Transformer-based architecture \cite{vaswani2017attention} that introduces several innovations for fusing visual and language data. The core components (summary in Figure \ref{fig:groundingdinoarchitechture}) include the following: (1) Visual Backbone through the integration of convolutional backbones to extract visual features from the input image, similarly to standard object detection models. These features are then embedded into the Transformer architecture. (2) Language Integration is a critical feature of Grounding DINO incorporating natural language descriptions for object identification similarly to open-world object detection \cite{joseph2021openworldobjectdetection}. To achieve this, it uses a pre-trained language model to encode textual inputs, such as category names or phrases. The encoded language vectors are then tightly integrated with visual features through a multi-step fusion process. (3) Query Selection and Cross-Modality Decoder: The model employs a feature enhancer that refines visual features using language input to guide the search for relevant objects. A language-guided query selection mechanism enables the identification of regions of interest in the image by selecting queries that match linguistic descriptions. Finally, a cross-modality decoder integrates these selected queries, refining detection and enabling accurate bounding box predictions and classifications even for previously unseen categories.

Grounding DINO's evaluation showcases its capabilities in both conventional and zero-shot detection tasks.
It achieves 52.5 average precision (AP) on the COCO zero-shot benchmark \cite{lin2015microsoftcococommonobjects}, indicating its strong ability to generalize without training data from the dataset.
The model sets a new standard on the ODinW zero-shot benchmark \cite{li2022groundedlanguageimagepretraining}, obtaining a mean 26.1 AP for open-world detection scenarios where the model must handle unseen object categories.
Grounding DINO also demonstrates strong performance on referring expression comprehension, meaning it can accurately detect objects when prompted by attribute-based descriptions, as seen in tasks like RefCOCO and RefCOCOg \cite{chen2024revisitingreferringexpressioncomprehension}.

%%%%%%%%%%%%%%%%%%%%%%%%%%%%%%%%%%%%%%%%%%
\section{Method: Aircraft Detection}\label{sec:result1_bench}
Typically, novel methods are assessed, validated, and contrasted with other algorithms to demonstrate their effectiveness. However, these methods are primarily evaluated on the COCO dataset, which is a comprehensive yet highly effective performance evaluator due to the number of images with multiple classes and resolutions. In this research, the main focus is evaluating these algorithms for aircraft detection from remote sensing, where the challenges are significantly different, and other types of noise and object sizes must be dealt with.

\begin{figure}[h]
        \centering
        \includegraphics[width=120mm,scale=1]{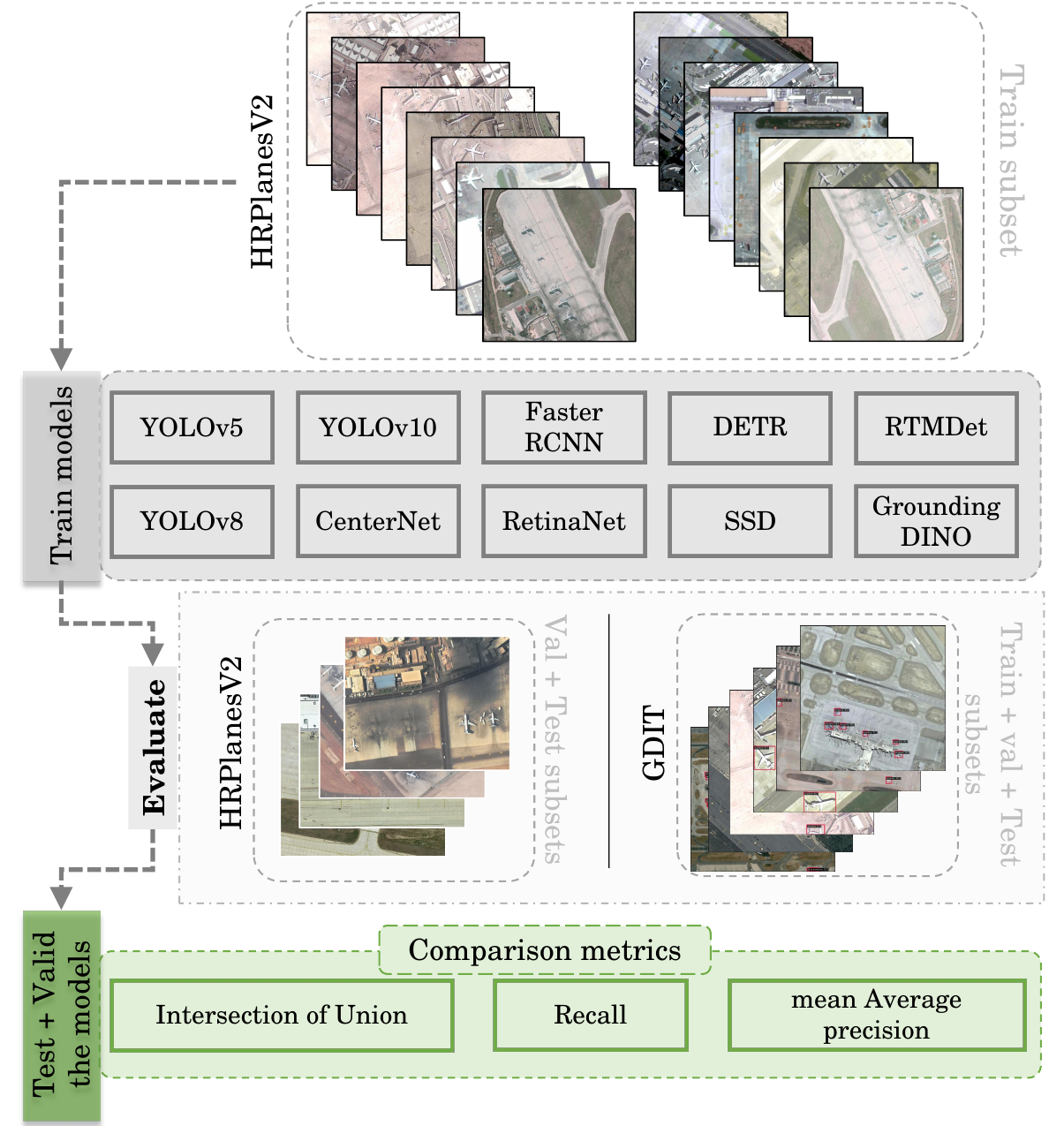}
        \caption{Flowchart of the FlightScope comparative study: The training is performed on HRPlanesV2 dataset and the Validation and Test conducted on HRPlanesV2 and GDIT Aerial airport datasets.}
        \label{fig:summary_HRP_GDIT_}
    \end{figure}

For the comparative study, the google earth \textbf{HRPlanesV2} dataset was selected as a training dataset because it contains the highest number of high resolution images (2120) among the other datasets (Table \ref{tab:dataset_overview}) while also providing different positions orientations and ground cases to the model for its generalization. As for the validation and test, unseen image from the HRPlanesV2 as well as the the GDIT Aerial airports dataset was included to evaluate with more accuracy the pre-trained models (Graphical summary of the work in the Fig. \ref{fig:summary_HRP_GDIT_}). The frameworks used in this comparative study are the one implemented by Ultralytics \cite{yolov8_ultralytics}, Detectron2 \cite{wu2019detectron2} and MMLab detection toolbox \cite{chen2019mmdetection}.

\subsection{Setup and Data Preparation}
\begin{figure}[h]
        \centering
        \includegraphics[width=\textwidth,scale=1]{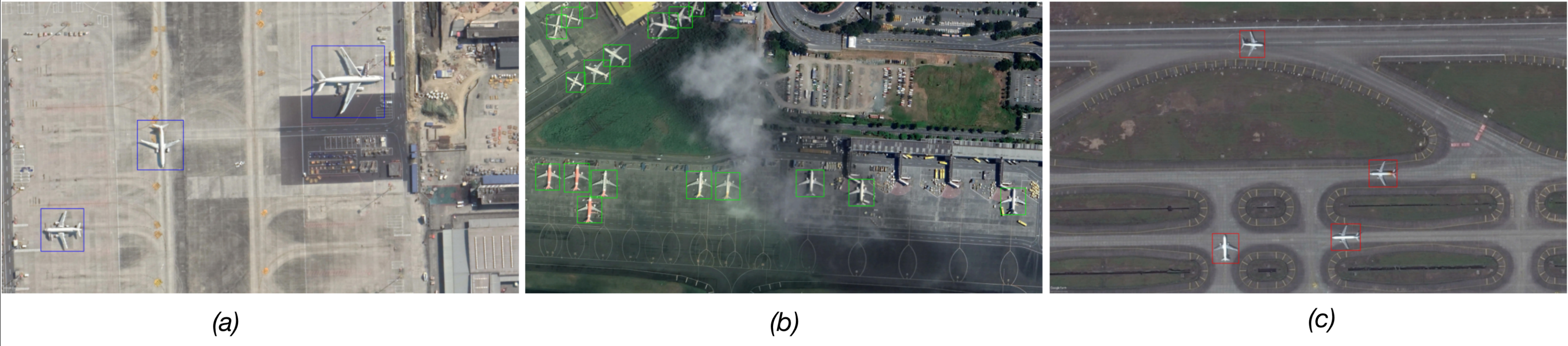}
        \caption{Sample preview from \textbf{HRPlanesV2} dataset: \textit{(a)} blue: training subset, \textit{(b)} green: test subset, \textit{(c)} red: validation subset}
        \label{fig:sample_preview_hrplanesv2}
    \end{figure}

Google earth \textbf{HRPlanesV2} is already annotated and splited into three subsets: 70\% for
training, 20\% for validation, and 10\% for testing but initially available in YOLO format while MMLab requires the dataset annotation format to be in COCO. Fig. \ref{fig:sample_preview_hrplanesv2} provide an sample from \textbf{HRPlanesV2} dataset with the corresponding annotation bounding boxes.

An overview of the setup configuration of the used training package is presented in Table \ref{config-setup}. For the training and the evaluation of the model 3 NVIDIA RTX A6000 each with 48G of memory have been used. These processors were made available by the \href{https://www.toelt.ai/}{TOELT LLC AI Lab}. The maximum number of epochs have been fixed to 500 and batch sizes to 32 and for some neural network architectures to 64 because the GPU memory allowed it.

\begin{table}[h]
\caption{\label{config-setup}System Configuration Setup}
\renewcommand{\arraystretch}{1.2}
\begin{tabular*}{\textwidth}{@{\extracolsep\fill}ll}
\toprule
\multicolumn{2}{c}{\textbf{Software Setup}} \\
\midrule
Name & Version  \\
\midrule
Ubuntu & 20.04.1 LTS \\
Python & 3.8 \\
PyTorch & 2.0.1  \\
CUDA & 12.2 \\
\midrule
\multicolumn{2}{c}{\textbf{Hardware Setup}} \\
\midrule
GPU & NVIDIA RTX A6000 48 GB $\times$ 3 \\
CPU & Intel(R) Core(TM) i9-10980XE CPU @ 3.00GHz \\
Memory & 128 GB \\
\bottomrule
\end{tabular*}
\end{table}

%\subsection{Evaluation Metrics}
%Three key metrics were employed to quantify the performance of the object detection models:

%\begin{itemize}
%  \item \textbf{Average Precision (AP):} A metric that measures the accuracy of object detection by evaluating the precision-recall curve, providing an overall assessment of detection performance.
%  \item \textbf{Recall:} This metric quantifies the ability of the model to correctly identify all relevant instances of objects, reflecting its sensitivity to detecting true positives.
%  \item \textbf{Intersection over Union (IoU):} IoU measures the spatial overlap between the predicted and ground truth bounding boxes, offering insights into the localization accuracy of detected objects.
%\end{itemize}

%The use of these metrics in the evaluation process ensures a nuanced understanding of the object detection models performance on the \textbf{GDIT} dataset, enhancing the reliability of the results across different subsets. 

\subsection{Evaluation Metrics}
Before delving into the specific evaluation metrics employed, it's essential to establish a comprehensive understanding of each metric's role in assessing the performance of object detection models, particularly concerning bounding box estimation accuracy. Within this study, PASCAL VOC metrics have been selected: AP, Recall and bounding boxes IoU \cite{padilla2021comparative}. 

\textbf{Average Precision (AP)} stands as a fundamental metric in object detection evaluation, providing a comprehensive measure of the model's ability to precisely identify objects of interest within an image. The AP metric, formulated in Eq. \ref{eq:aver_prec}, is computed by integrating the precision-recall curve $p(r)$, which represents the trade-off between true positive detections and false positives across various confidence thresholds. This integration yields a scalar value reflecting the model's overall detection accuracy, with higher AP scores indicating superior performance.

\begin{equation}
    \text{AP} = \int_{0}^{1} p(r) \, dr
    \label{eq:aver_prec}
\end{equation}

\textbf{Recall}, another metric in object detection assessment, quantifies the model's ability to correctly identify all relevant instances of objects present within an image. It signifies the sensitivity of the model in capturing true positives (TP) while minimizing false negatives (FN). Mathematically, Recall is defined as the ratio of true positive detections to the total number of ground truth objects (Eq. \ref{eq:recall}).

\begin{equation}
    \text{Recall} = \frac{\text{TP}}{\text{TP} + \text{FN}}
    \label{eq:recall}
\end{equation}

Lastly, \textbf{Intersection over Union (IoU)} serves as a metric for evaluating the spatial alignment between predicted bounding boxes and ground truth annotations. IoU quantifies the extent of overlap between these bounding boxes, providing insight into the localization accuracy of detected objects. IoU, expressed by Eq. \ref{eq:IoU}, is computed as the ratio of the intersection area between the predicted and ground-truth bounding boxes to their union.

\begin{equation}
    \text{IoU} = \frac{\text{Area of Intersection}}{\text{Area of Union}}
    \label{eq:IoU}
\end{equation}

Employing these metrics in the evaluation process enables an accurate assessment of object detection model performance, particularly concerning bounding-box estimation accuracy.

\subsection{Results}

\begin{figure}[H]
        \centering
        \includegraphics[width=140mm,scale=1]{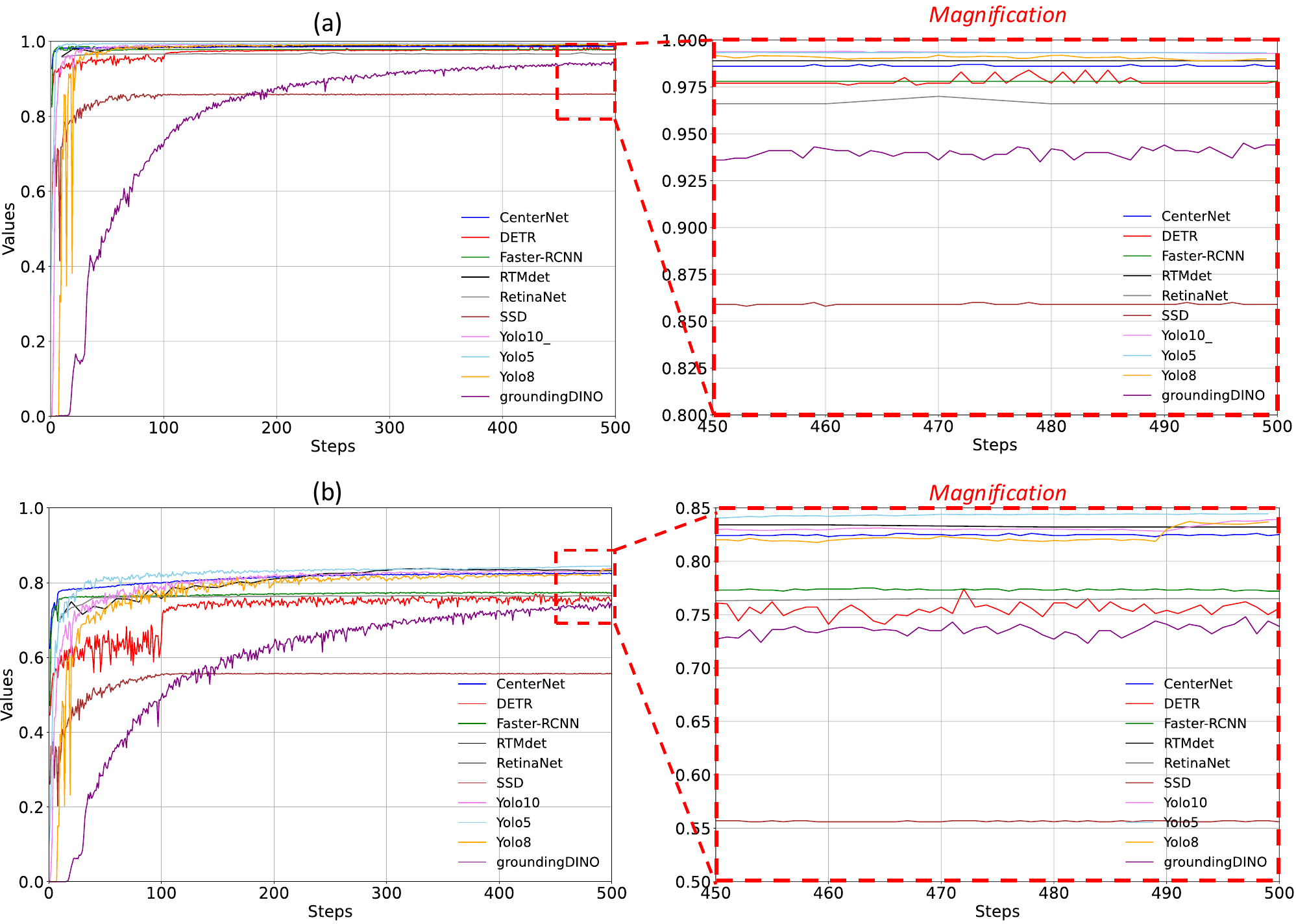}
        \caption{Comparison of bounding box mean average precision (mAP) curves for trained object detection algorithms. Raw figures of the curves are on the left; the right figures are magnifications from epochs 450 to 500. %EE: check meaning retained
        		{(\textbf{a})} represents the mAP. {(\textbf{b})} illustrates the mAP50.}
        \label{fig:overall_map_curves}
  
\end{figure}

This subsection presents a detailed result showcase of the object detection algorithm training during the 500 epochs. Figure \ref{fig:overall_map_curves} displays the mAP (overall mean average precision %EE: check meaning retained
across different confidence thresholds) and the mAP50 (for objects detected by an intersection over union (IoU) threshold of 50\% and up) curves collectively visualizing the performance of the object detection algorithms in the task of aircraft detection from remote sensing imagery. Notably, the performances of the algorithms are quite comparable with an overall mAP varying between 0.86 and 0.99. Among the models, YOLOv5 emerges as a standout performer, achieving the highest mAP value of 0.99471 at step 150, showcasing its great precision and robustness. YOLOv8 and YOLOv10 closely follow, both reaching a peak mAP value of about 0.98, emphasizing the efficacy of the YOLO architecture in aerial object detection. However, SSD lags behind with a comparatively modest mAP value of 0.86 at step 74, which stabilizes in the rest of the training process.

The results in Figure \ref{fig:overall_map_curves}b confirm the previous discussion, as YOLOv5 continues to outperform, achieving the highest mAP50 value of 0.84454 at step 493. RTMDet and YOLOv8 remain strong with mAP50 values of 0.838 at step 340 and 0.8372 at step 492, respectively. %EE: check meaning retained
 CenterNet consistently performs well, achieving an mAP50 value of 0.826 at step 439. Faster RCNN maintains a balanced mAP50 value of 0.775 at step 402, showcasing reliability in detection. DETR contributes robustly with an mAP50 value of 0.774 at step 472, while RetinaNet exhibits stability with an mAP50 value of 0.765 at step 250. Finally, SSD, with a noticeable gap between mAP and mAP50 values, suggests potential challenges in localization precision, emphasizing the need for refinement in specific scenarios.

The results in Fig. \ref{fig:overall_map_curves}-b confirm the previous discussion, as YOLOv5 continues to outperform, achieving the highest mAP50 value of 0.84454 at step 493. RTMDet and YOLOv8 remains strong with mAP50 values of 0.838 at step 340 and 0.8372 at step 492 successively. CenterNet consistently performs well, achieving an mAP50 value of 0.826 at step 439. Faster-RCNN maintains a balanced mAP50 value of 0.775 at step 402, showcasing reliability in detection. DETR contributes robustly with an mAP50 value of 0.774 at step 472, while RetinaNet exhibits stability with an mAP50 value of 0.765 at step 250. Finally, SSD, with a noticeable gap between mAP and mAP50 values, suggests potential challenges in localization precision, emphasizing the need for refinement in specific scenarios.

%\begin{figure}[h!]
%        \centering
%        \includegraphics[width=\textwidth,scale=1]{bboxloss.pdf}
%        \caption{Bounding box loss curves of the trained object detection algorithms.}
%        \label{fig:bboxloss}
%\end{figure}
%\begin{figure}[h]
%        \centering
%        \includegraphics[width=\textwidth,scale=1]{totalloss.pdf}
%        \caption{Total loss curves of the trained object detection algorithms.}
%        \label{fig:totalloss}
%\end{figure}

\begin{figure}[h!]
    \centering
    \begin{subfigure}{0.49\textwidth} % Adjust the width as needed
        \centering
        \includegraphics[width=\textwidth]{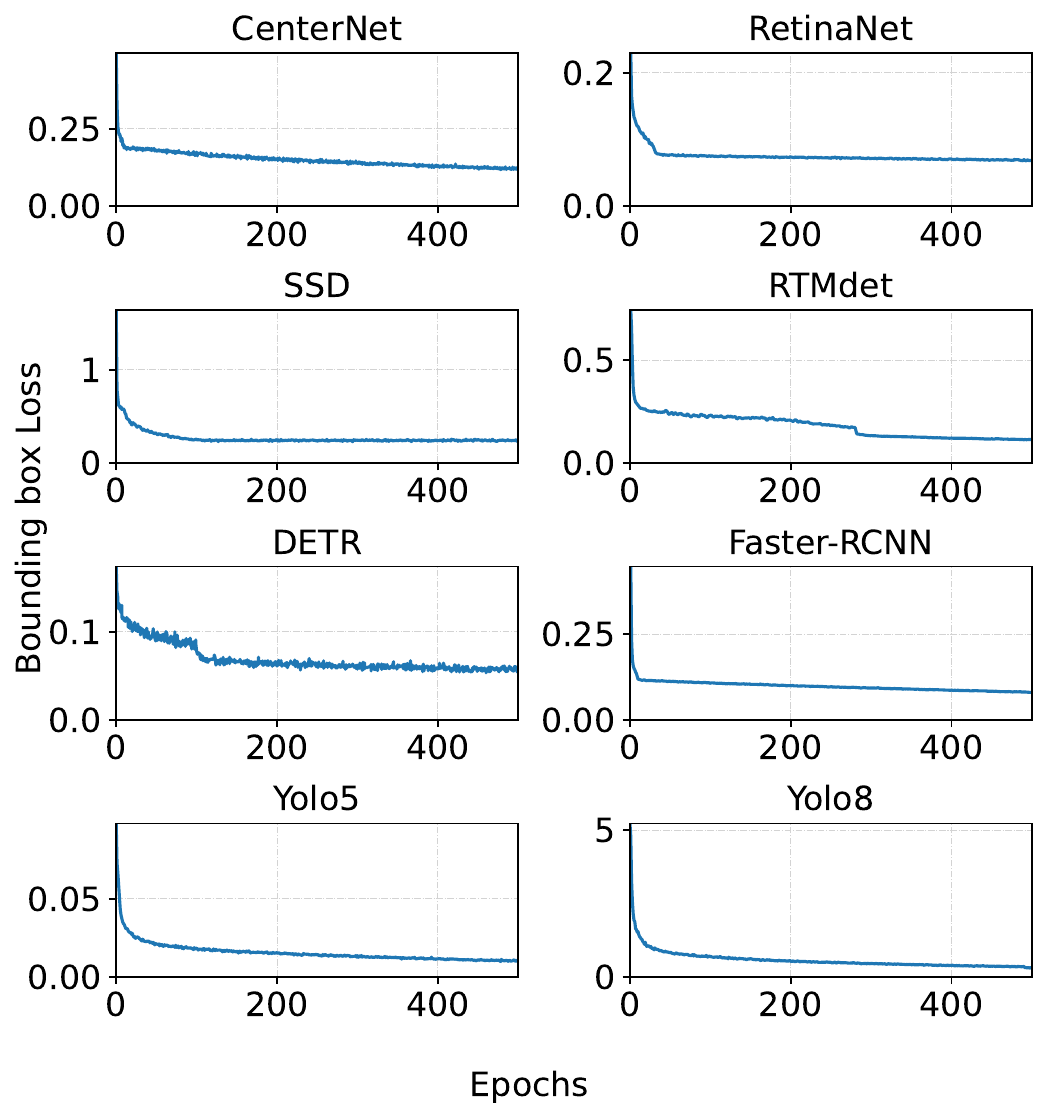}
        \caption{Bounding box loss}
        \label{fig:bboxloss}
    \end{subfigure}
    \hfill
    \begin{subfigure}{0.49\textwidth} % Adjust the width as needed
        \centering
        \includegraphics[width=\textwidth]{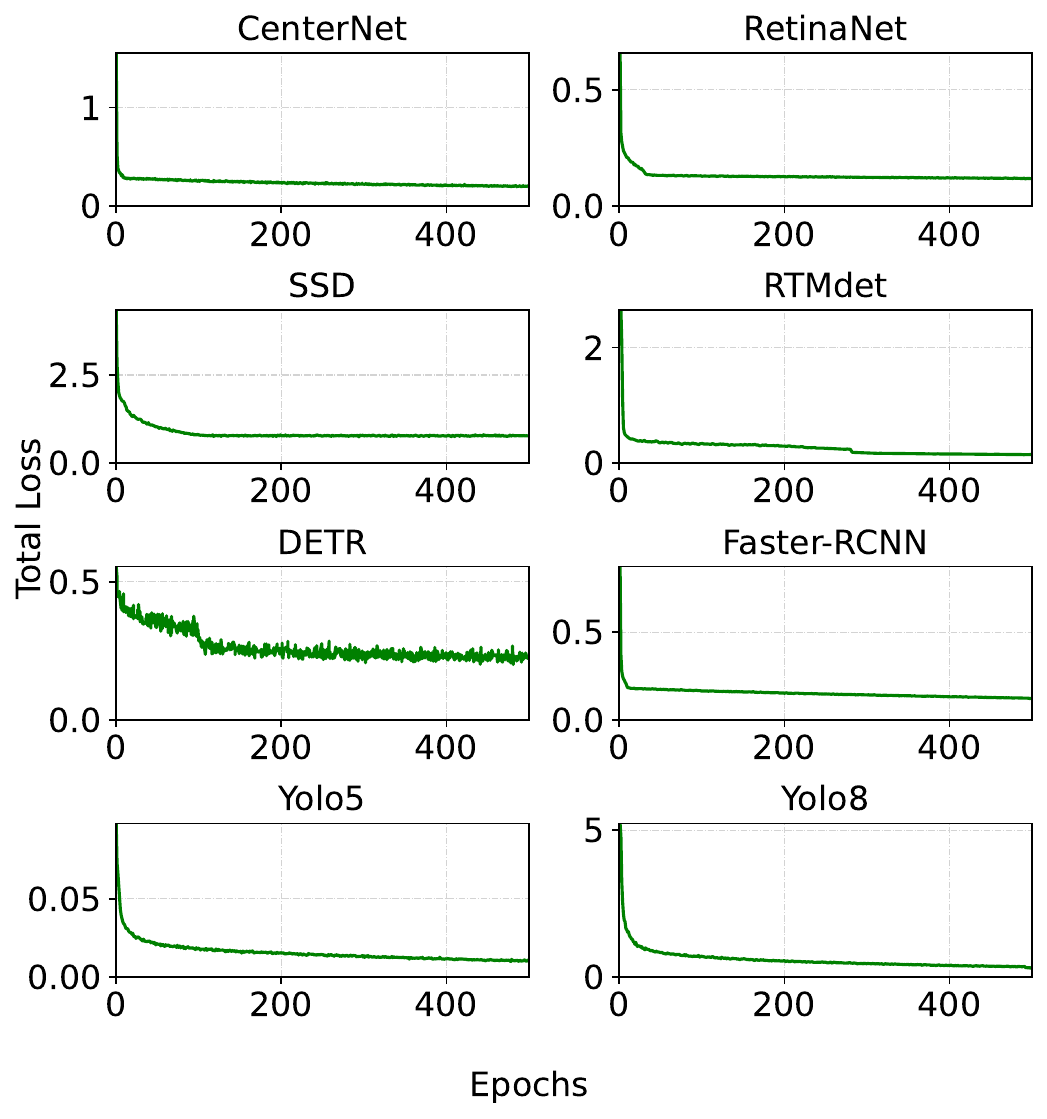}
        \caption{Total loss}
        \label{fig:totalloss}
    \end{subfigure}
    \caption{Comparison of loss curves for the 8 trained object detection models: (a) Bounding box loss curves (b) Total loss curves.}
    \label{fig:loss_comparison}
\end{figure}

In addition, Fig. \ref{fig:bboxloss} and \ref{fig:totalloss} provide successively a visualization of the bounding box and total loss curves for the algorithms, where it is noticeable that the majority of the algorithm converge quickly around 100 epochs (apart from RTMDet which has a jump value around the epoch 280) and reach a seemingly horizontal asymptote line within 300 epochs. While extending the training duration beyond 500 epochs might promote further convergence, there is a potential risk of overfitting the models.

\subsection{Evaluation on GDIT Dataset}

\begin{figure}[H]
     
        \includegraphics[width=110mm,scale=1]{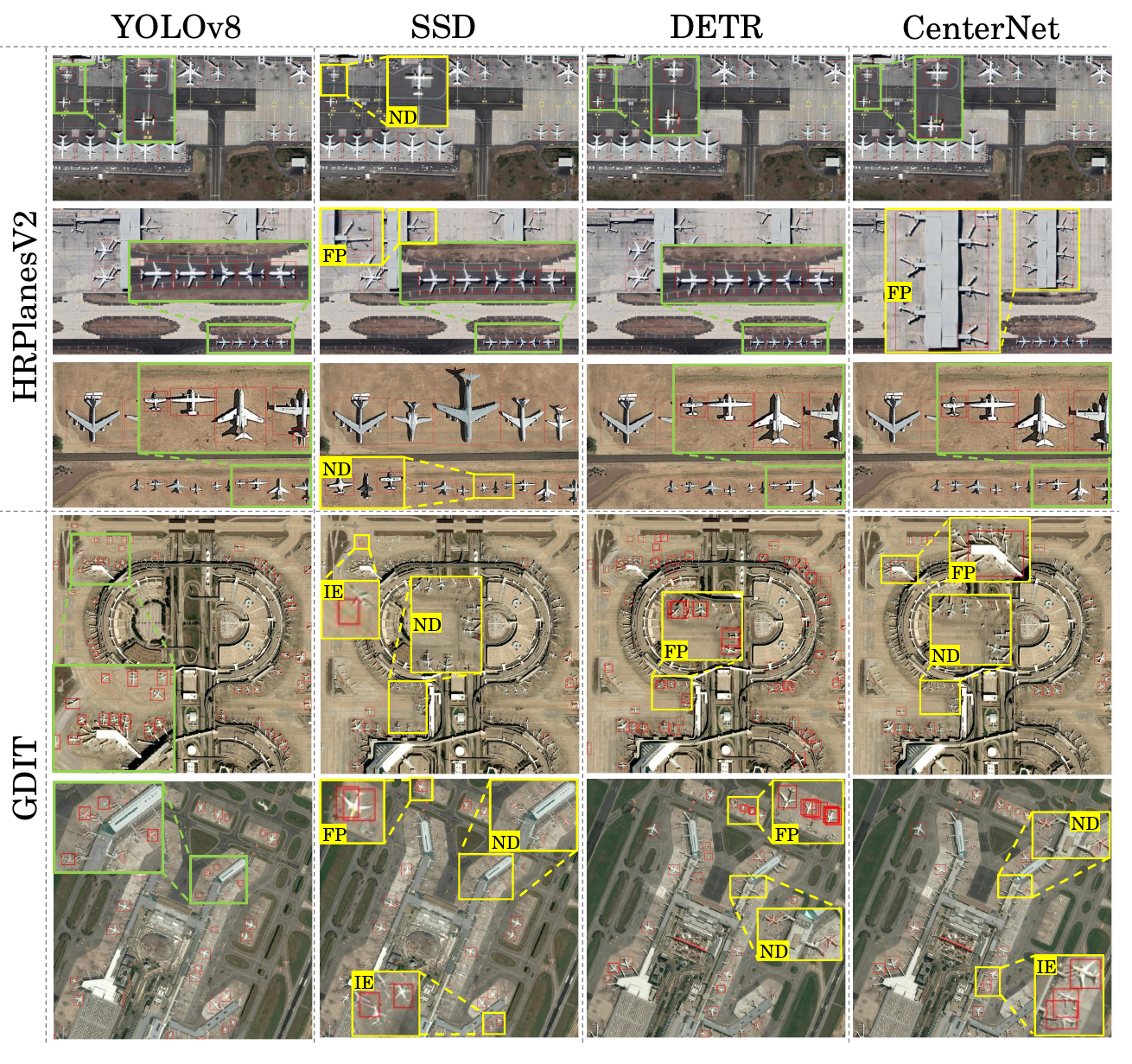}
        \caption{Inference examples of YOLOv8, DETR, SSD, and CenterNet on unseen images from the Google Earth HRPlanesv2 dataset and Airbus GDIT. Green boxes showcase the accurate detections, `FP' stands for "false positive", `ND' for "no detection" and `IE' for "inaccurate estimation".
}
        \label{fig:test_GDIT_HRPlanes1}
\end{figure}

\begin{figure}[H]
       
        \includegraphics[width=110mm,scale=1]{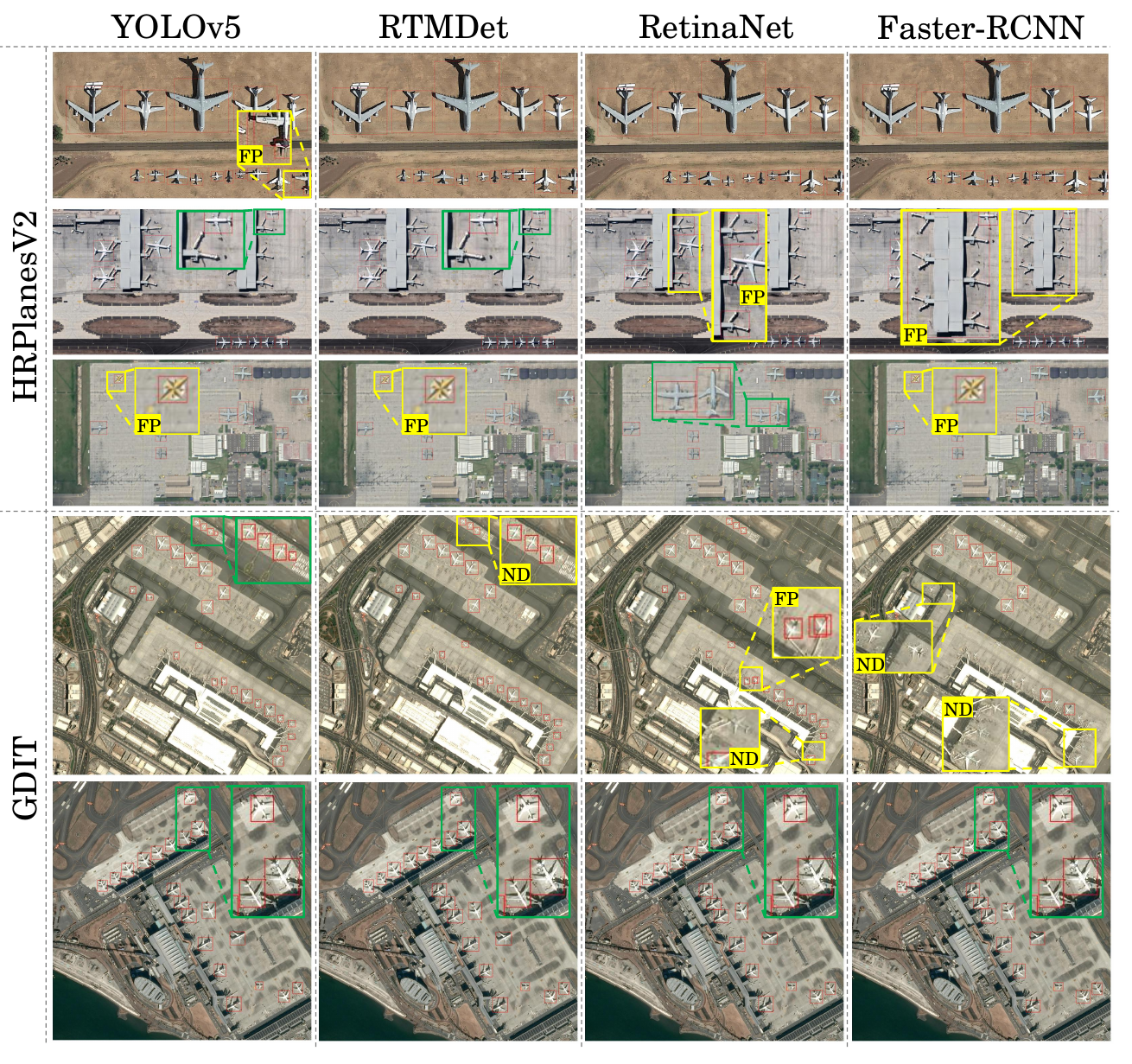}
        \caption{Inference examples of YOLOv5, RTMDet, RetinaNet and Faster-RCNN on another set of unseen images from Google Earth HRPlanesv2 dataset and Airbus GDIT. Green boxes showcase the accurate detections, `FP' stands for "false positive", `ND' for "no detection".
}
        \label{fig:test_GDIT_HRPlanes2}
\end{figure}

To evaluate the efficacy of object detection models, originally trained on the HRPlanesv2 dataset, a thorough assessment was conducted using the GDIT dataset. The evaluation encompassed all subsets of images: train, test, and validation, each representing diverse scenarios. This comprehensive evaluation aimed to gauge the adaptability and robustness of the algorithms under varied conditions encountered in different subsets. A sample of unseen images from both datasets highlighting the bounding boxes of estimated aircraft is presented in Figures \ref{fig:test_GDIT_HRPlanes1} and \ref{fig:test_GDIT_HRPlanes2} where ND, FP, and IE stand for `Not Detected', `False Positives', and `Inaccurate Estimation', respectively. The figures show the struggle of CenterNet, Faster RCNN, and SSD in the detection of small objects, while both YOLO versions and RTMDet are able to detect over 80\% of the aircraft in the image with a minimum confidence of 32.6\% with small amounts of FPs and/or ND.

\begin{figure}[H]
        \centering
        \includegraphics[width=\textwidth,scale=1]{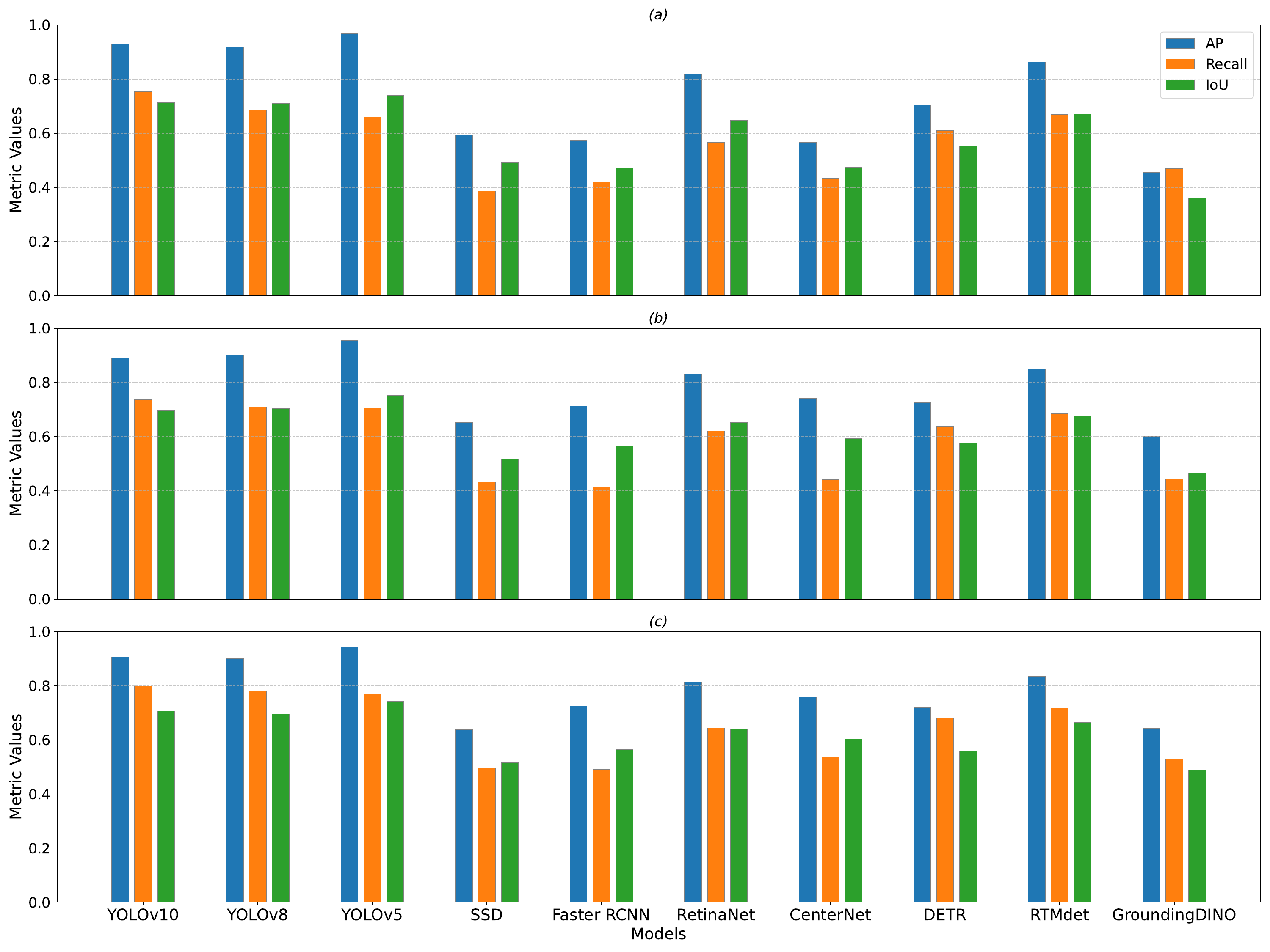}
        \caption{Estimated evaluation metrics when inferencing the 8 models (initially trained on HRPlanesV2) on all images from unseen subsets `Train' {(\textbf{a})}, `Test' {(\textbf{b})} and `Validation' {(\textbf{c})} from the GDIT aircraft dataset.}
        \label{fig:Metricscomparision}

\end{figure}

The results of the metrics IoU, recall, and AP are presented in histograms (Figure \ref{fig:Metricscomparision}). On the training subset, YOLOv8 demonstrated notable performance with an AP of 91.9\%, recall of 68.6\%, and IoU of 71.1\%. YOLOv5 exhibited excellence with an AP of 96.8\%, recall of 66.1\%, and IoU of 74.0\%. Conversely, SSD displayed a comparatively lower AP of 59.4\%, recall of 38.6\%, and IoU of 49.2\%. Performance metrics such as AP, recall, and IoU showed variations for Faster RCNN and CenterNet.

The evaluation on the test subset provided further insights into the generalization capabilities of the pre-trained algorithms. YOLOv8 maintained high performance with an AP of 90.3\%, recall of 70.9\%, and IoU of 70.4\%. Similarly, YOLOv5 exhibited commendable performance with an AP of 95.6\%, recall of 70.5\%, and IoU of 75.2\%. However, SSD, Faster RCNN, and CenterNet displayed variations in performance metrics across the test subset.

\vspace{-6pt}

Finally, in the validation subset, distinct challenges were encountered, revealing the algorithms' robustness in diverse scenarios. YOLOv8 achieved an AP of 90.0\%, recall of 78.1\%, and IoU of 69.6\%. YOLOv5 maintained high standards with an AP of 94.2\%, recall of 77.0\%, and IoU of 74.2\%. Performance nuances were observed for SSD, Faster RCNN, and CenterNet, underscoring their adaptability to distinct datasets.

Table \ref{tab:object_detection_performance} summarizes the object detection performance metrics of various models on remote sensing images across different subsets—train, test, validation, and all the datasets. {Furthermore, as expected, the YOLO  models stand out as the top real-time prediction models, with YOLOv10 being the best with 54.3 frames per second (FPS) due to its reduced number of trainable parameters. The overall performances showed that YOLOv5 emerged as the top-performing algorithm across all subsets in terms of accuracy, with the highest evaluation metrics AP and IoU. Notably, YOLOv5 demonstrated commendable recall rates and IoU scores, positioning it as the leading algorithm for aircraft detection in this study. Additionally, YOLOv5 also yielded a real-time performance of 37.4 FPS, usually acceptable in industrial environments.}
Conversely, SSD consistently exhibited comparatively lower performance metrics, indicating challenges in accurately detecting aircraft instances across subsets. While other algorithms, including YOLOv8, Faster RCNN, and CenterNet, displayed varying degrees of success, YOLOv5 consistently outperformed them in terms of AP, recall, and IoU.

\begin{table}[h!]
    \caption{Trained models for aircraft detection and classification performance on remote sensing images. Tr. subset refers to all images from `train', Te. subset refers to all images from `test', Val. subset refers to all images from `validation', and All dataset refers to all images in the dataset.}\label{tab:object_detection_performance}
   
    \begin{tabular*}{\textwidth}{@{\extracolsep{\fill}}lcccccc}
    \toprule
    \textbf{Architecture} & \textbf{Real-Time Perf.} & \textbf{Metric} & \textbf{Tr. Subset} & \textbf{Te. Subset} & \textbf{Val. Subset} & \textbf{All Dataset} \\
    \midrule
    YOLOv8 & 43.4 FPS & AP & 0.919 & 0.903 & 0.900 & 0.907 \\
           &          & Recall & 0.686 & 0.710 & 0.781 & 0.726 \\
           &          & IoU & 0.711 & 0.704 & 0.696 & 0.704 \\
    \midrule
    YOLOv10 & {\textbf{54.3 FPS}} & {AP} & 0.929 & 0.890 & 0.906 & 0.908 \\
           & {}          & {Recall} & {0.754} & 0.736 & 0.799 & 0.763 \\
           & {}          & {IoU} & 0.713 & 0.696 & 0.706 & 0.705 \\
    \midrule
    YOLOv5 & 37.4 FPS & AP & 0.968 & 0.956 & 0.942 & \textbf{0.955} \\
           &          & Recall & 0.661 & 0.705 & 0.770 & 0.712 \\
           &          & IoU & 0.740 & 0.752 & 0.742 & 0.745 \\
    \midrule
    SSD & 9.7 FPS & AP & 0.594 & 0.653 & 0.638 & 0.628 \\
        &         & Recall & 0.386 & 0.432 & 0.496 & 0.438 \\
        &         & IoU & 0.492 & 0.517 & 0.516 & 0.508 \\
    \midrule
    Faster RCNN & {7.5 FPS} & AP & 0.573 & 0.714 & 0.726 & 0.671 \\
                &         & Recall & 0.422 & 0.413 & 0.491 & 0.442 \\
                &         & IoU & 0.473 & 0.564 & 0.564 & 0.533 \\
    \midrule
    RetinaNet & {6.6 FPS} & AP & 0.819 & 0.830 & 0.815 & 0.821 \\
              &         & Recall & 0.566 & 0.620 & 0.644 & 0.610 \\
              &         & IoU & 0.647 & 0.652 & 0.641 & 0.647 \\
    \midrule
    CenterNet & {8.5 FPS} & AP & 0.566 & 0.742 & 0.758 & 0.689 \\
              &         & Recall & 0.434 & 0.441 & 0.536 & 0.470 \\
              &         & IoU & 0.473 & 0.593 & 0.603 & 0.556 \\
    \midrule
    DETR & {6.1 FPS} & AP & 0.705 & 0.725 & 0.718 & 0.716 \\
         &         & Recall & 0.610 & 0.636 & 0.680 & 0.642 \\
         &         & IoU & 0.553 & 0.576 & 0.558 & 0.563 \\
    \midrule
    RTMDet & 9.2 FPS & AP & 0.863 & 0.850 & 0.836 & 0.850 \\
           &         & Recall & 0.670 & 0.685 & 0.718 & 0.691 \\
           &         & IoU & 0.672 & 0.676 & 0.664 & 0.670 \\
    \midrule
    {GroundingDINO} & {3.2 FPS} & {AP} & 0.456 & 0.601 & 0.641 & 0.566 \\
           & {}          & {Recall} & 0.467 & 0.445 & 0.531 & 0.481 \\
           & {}          & IoU & 0.362 & 0.466 & 0.488 & 0.439 \\
    \bottomrule
\end{tabular*}

\end{table}

Table \ref{tab:object_detection_performance} summarizes the object detection performance metrics of various models on remote sensing images across different subsets—Train, Test, Validation and all the dataset. The overall performances showed the YOLOv5 emerged as the top-performing algorithm across all subsets, with the highest evaluation metrics AP and IoU. Notably, YOLOv5 demonstrated commendable recall rates and IoU scores, positioning it as the leading algorithm for aircraft detection in this study.
Conversely, SSD consistently exhibited comparatively lower performance metrics, indicating challenges in accurately detecting aircraft instances across subsets. While other algorithms, including YOLOv8, Faster RCNN, and CenterNet, displayed varying degrees of success, YOLOv5 consistently outperformed them in terms of AP, Recall, and IoU.

\newpage
%%%%%%%%%%%%%%%%%%%%%%%%%%%%%%%%%%%%%%%%%%%%%%%%
%%%%%%%%%%%%%%%%%%% SECTION 6 %%%%%%%%%%%%%%%%%%
%%%%%%%%%%%%%%%%%%%%%%%%%%%%%%%%%%%%%%%%%%%%%%%%

\section{Conclusion}\label{sec:conclusion}
This study presents a comprehensive evaluation of aircraft detection algorithms in satellite imagery, namely YOLO (v5 and v8), Faster RCNN, CenterNet, RetinaNet, RTMDet and DETR, specifically focusing on the HRPlanesV2 dataset and extending the assessment to subsets of the GDIT dataset: train, test, and validation. The used training setup involves training the aforementioned object detection algorithms on the dataset for 500 epochs on three NVIDIA RTX A6000 GPUs, with each GPU having 48 GB of memory. %EE: check meaning retained
The results of the evaluation demonstrate the adaptability and robustness of the trained object detection algorithms. Among these algorithms, YOLOv5 emerges as the standout performer, achieving the highest mean average precision (mAP) of 0.99, highlighting its precision and robustness. YOLOv8 and v10 closely follow, further emphasizing the effectiveness of the YOLO architecture in aerial object detection. On the other side, SSD displayed the lowest performances in both the training and evaluation.
Furthermore, the evaluation extends to the GDIT dataset, providing a more comprehensive assessment by deploying the trained network on other scenarios, including different satellite imagery sources. By employing evaluation metrics such as average precision, recall, and intersection over union, YOLOv5 still consistently outperforms the other algorithms, demonstrating superior performance across all subsets. This solidifies YOLOv5's position as the top-performing algorithm for aircraft detection, characterized by obtained AP, recall, and IoU scores.

The topic of this research offers an extensive overview and comparative study, providing an in-depth analysis of the performance, accuracy, and computational demands of object detection methods. The insights gained from this study significantly enhance the decision-making process for selecting the most effective aircraft localization techniques in satellite imagery, supported by detailed training and validation processes using the HRPlanesv2 dataset and additional validation with the GDIT dataset. 

%\section*{Acknowledgments}
%This was was supported in part by......

%Bibliography
\bibliographystyle{unsrt} %elsarticle-harv
\bibliography{references}

\end{document}